\documentclass{article}

\usepackage[accepted]{icml2026}

\usepackage{hyperref}
\usepackage{url}

\usepackage{packages}

\usepackage{amsmath,amsfonts,bm}
\usepackage{xurl}

\newtheorem{theorem}{Theorem}[section]

\theoremstyle{definition}
\newtheorem{definition}[theorem]{Definition}

\theoremstyle{remark}

\def\eqref#1{Eq.~\ref{#1}}

\def\1{\bm{1}}

\def\eps{{\epsilon}}

\DeclareMathAlphabet{\mathsfit}{\encodingdefault}{\sfdefault}{m}{sl}
\SetMathAlphabet{\mathsfit}{bold}{\encodingdefault}{\sfdefault}{bx}{n}

\newcommand{\R}{\mathbb{R}}

\renewcommand{\eps}{\varepsilon}

\renewcommand{\cal}{\mathcal}

\newcommand{\priv}{\ensuremath{\mathrm{priv}}}
\newcommand{\pub}{\ensuremath{\mathrm{pub}}}

\icmltitlerunning{Membership Inference Attacks for Unseen Classes}

\begin{document}

\twocolumn[
  \icmltitle{Membership Inference Attacks \\ for Unseen Classes}



  \icmlsetsymbol{equal}{*}

  \begin{icmlauthorlist}
    \icmlauthor{Pratiksha Thaker}{equal,cmu}
    \icmlauthor{Neil Kale}{equal,cmu}
    \icmlauthor{Zhiwei Steven Wu}{cmu}
    \icmlauthor{Virginia Smith}{cmu}
  \end{icmlauthorlist}

  \icmlaffiliation{cmu}{Carnegie Mellon University}

  \icmlcorrespondingauthor{Pratiksha Thaker}{pthaker@andrew.cmu.edu}
  \icmlcorrespondingauthor{Virginia Smith}{smithv@cmu.edu}

  \icmlkeywords{Machine Learning, ICML}

  \vskip 0.3in
]



\printAffiliationsAndNotice{}  

\begin{abstract}

A key tool in developing safe AI models is \emph{data auditing},
i.e., using statistical tools to determine whether harmful content
may have been used in the training data of a black-box model.
Unfortunately, most \emph{membership inference attacks} (MIAs)
used to perform this type of auditing themselves assume
\emph{access} to examples of harmful content from the same distribution as the query data.
In real-world auditing scenarios, auditors often face legal and ethical
restrictions preventing them from accessing a representative set of samples of harmful content to train MIA models effectively.
We abstract and formalize this setting into a new data access model,
the ``unseen class'' setting,
and show that the state of the art MIAs fail due to the lack of access to the full target distribution.
We show in this setting, \emph{quantile regression attacks}
outperform approaches typically considered to be SoTA. 
We demonstrate this both empirically and theoretically, showing that quantile regression attacks achieve up to \textbf{11$\times$ the TPR} of shadow model-based approaches in practice, and providing a theoretical model that outlines the generalization properties required for this approach to succeed. Our work identifies an important failure mode in existing MIAs and provides a cautionary tale for practitioners that aim to directly use existing tools for real-world applications of AI safety.

\end{abstract}

\section{Introduction}
\label{introduction}


The most effective membership inference attacks require training on data from the target class -- but what happens when accessing that data is illegal, unethical, or impossible?
For example, in the domain of child safety,
model providers, ML developers, and law enforcement are among many entities
who are invested in ensuring that models are not trained on child sexual abuse material (CSAM)~\citep{thiel2023identifying,thiel2023generative,kapoor2024societal,thornsafetybydesign} -- 
but legal access to this data even for safety purposes is, for good reason, extremely restricted.

A well-studied statistical tool for auditing training data
is a membership inference attack (MIA)~\cite{shokri2017membership, lira}.
The most common and state-of-the-art MIA approaches 
train a large number of \emph{shadow models}~\cite{lira} --
models that aim to mimic the behavior of the target model
but are trained on controlled subsets of data.
These attacks typically assume access to a
``background'' distribution of data sampled from a distribution
identical to the data used to train the target model,
but disjoint from the target model's training set.
But in real-world domains like child safety,
an identical background distribution is unavailable --
in fact, the target subpopulation of interest is exactly the one
that the auditor cannot access when training attacks.

\begin{figure}[t!]
        \centering
        \includegraphics[width=0.45\textwidth]{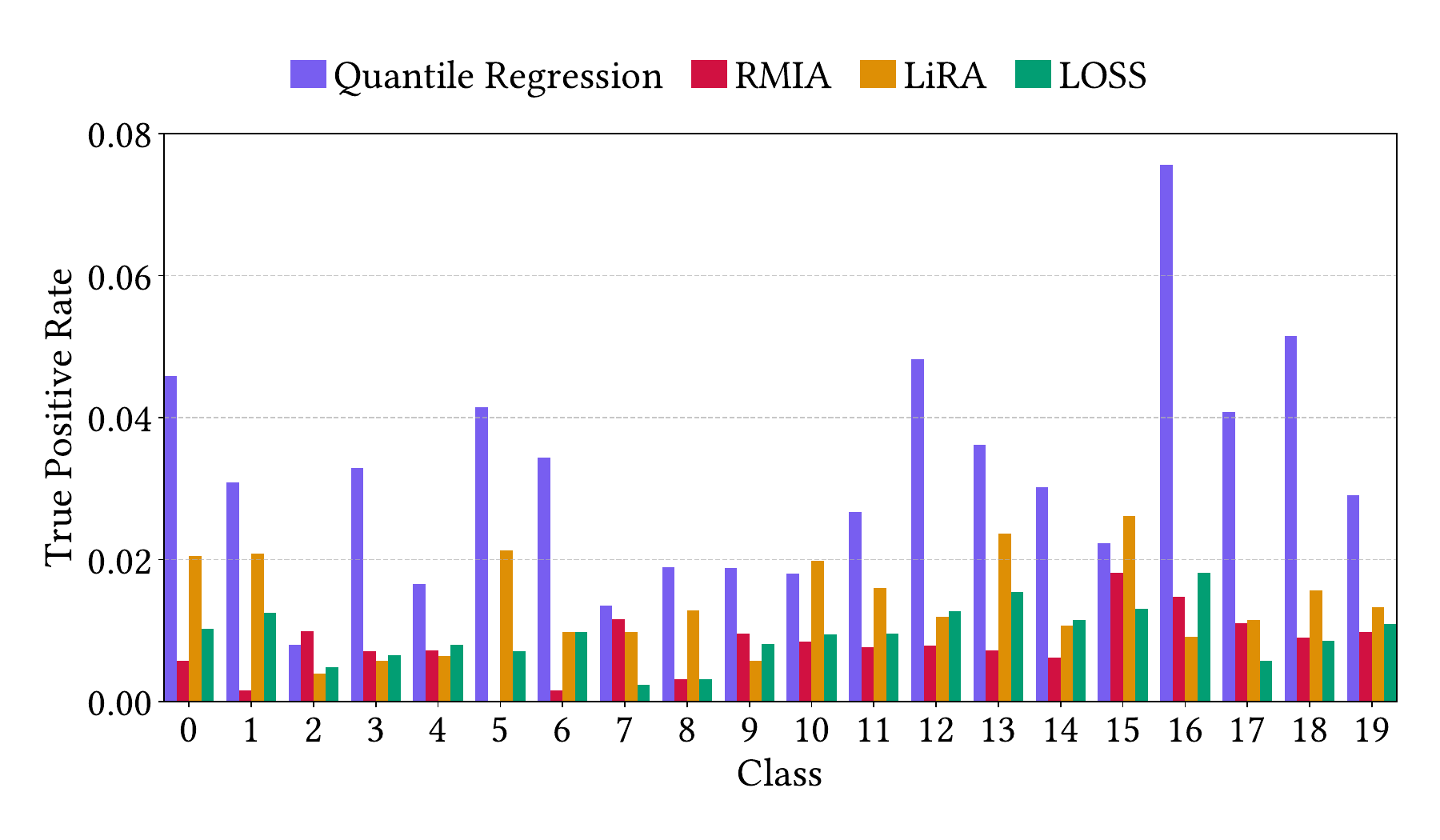} 
        \setlength{\belowcaptionskip}{-0.2in}
        \caption{True positive rates at 1\% false positive rate when each of 20 classes from CIFAR-100 (superclass set) are not seen at attack training time. Shadow model-based attacks struggle due to the lack of training samples, falling to near baseline performance (LOSS), while the quantile regression attack dramatically outperforms them by finding features that generalize \emph{across} classes, including unseen classes.}
        \label{fig:qr_cifar100_intro}
\end{figure}

This motivates what we call the unseen-class setting: membership inference where the auditor has no samples from the target class. In the setting of CSAM detection, auditors \emph{cannot} directly access examples of CSAM themselves to train proxy models for membership inference~\citep{thiel2023identifying} due to legal and ethical restrictions -- only law enforcement may have access to a few \emph{query} samples that cannot be used during training. Beyond CSAM, this captures auditing medical records, proprietary data, or any resource-constrained third-party audit. 
As a result, a fundamental assumption in shadow model-based MIAs is broken in this highly practical setting. We show that shadow model attacks -- widely considered the gold standard -- fail catastrophically in this setting, while quantile regression attacks achieve up to 11× higher TPR.

This example is part of a broader set of applications that we formalize in this work, in which an auditor or adversary wants to query membership of samples from some class $i$ in the data,
but has \emph{no samples} from class $i$ when training the attack.
While prior work has studied less extreme forms of distribution shift
(where all target classes are still present \cite{yichuan2024assessing, liu2022membership, lira, shokri2017membership}),
this is a new data access model not previously studied in the MIA literature, 
which we call the \textit{``unseen class''} setting.
The unseen-class setting abstracts the critical property of
the CSAM auditing scenario,
that examples from the target class are \emph{not available}
to the auditor when training the attack -- 
not even a small number of samples.

This setting captures not only the CSAM auditing scenario but also other safety- or resource-critical auditing scenarios: 
for example, where the target training data may be proprietary or sensitive (such as auditing models trained on medical records) 
or where attack training data may be scarce (e.g. if a third party is auditing a large company's model).

In this work, we show that, although shadow model-based attacks are considered state-of-the-art in the research literature, they \emph{fail catastrophically} when performing membership inference in the unseen class setting.
Shadow models crucially rely on training models that \emph{perform the same task as the target model}, 
and with no samples from class $i$, 
a classifier assigns zero probability to label $i$,
causing the classification task to fail on samples from the unseen class.
We are the first to identify a significant failure mode of shadow model-based attacks, 
which are generally considered to be the gold standard for MIA \citep{lira, zarifzadeh2023rmia}.

The failure of shadow model-based attacks shows that the unseen-class setting is challenging.
However, we identify that a simple, computationally efficient baseline -- \emph{quantile regression} attacks ~\citep{bertran2023scalable, tang2023membership} -- significantly outperforms shadow model attacks in this scenario.
Quantile regression attacks
were originally proposed as an MIA that can provide a provable guarantee on the attack's false-positive rate by predicting an $\alpha$-quantile of the nonmember score distribution.
We observe that quantile regression attacks also \emph{generalize} well to unseen classes,
because they learn features that correlate with score across classes.
This gives them an unexpected advantage in the unseen-class setting.

We verify this intuition both theoretically and empirically, using a set of benchmarks with image, language, and tabular data to investigate the unseen class setting in practice.

Overall, we make the following contributions:
\begin{itemize}[leftmargin=*]
\itemsep0.1em 
\item We identify a new MIA data access model, the unseen-class setting. This setting captures concerns in real-world auditing where data access is limited due to legal, ethical, or logistical constraints, such as CSAM detection, where the target data is not available when the attack models are being trained.
\item We show that, in this setting, the most popular and state-of-the-art approaches based on training shadow models deteriorate in performance -- on par with or worse than a simple baseline based on global thresholding. 
\item We evaluate a baseline based on \emph{quantile regression} attacks and find surprisingly strong results in the unseen-class setting across data domains: 
\begin{itemize}
\item On image data, in the 1\% FPR regime, quantile regression can achieve up to $\mathbf{11\times}$ the TPR of shadow models on the unseen class (on CIFAR-100).
Meanwhile, on ImageNet, we find that quantile regression can achieve 3.8\% TPR at 1\% FPR (about half the TPR achieved by full training) with access to only \emph{10\% of training classes.}
\item On tabular data, in the 1\% FPR regime, quantile regression achieves up to 2$\times$ the TPR of shadow models and improves AUC by 10 points.
\item Attacking a GPT-2 classifier, quantile regression achieves 6$\times$ the TPR of shadow models in text classification (20 Newsgroups) in the 1\% FPR regime.
\end{itemize}
\item Finally, we provide a theoretical model illustrating the benefits and potential limitations of quantile regression in this setting. Our analysis helps to better explain the effectiveness of the approach and also points to several directions of future study.
\end{itemize}

As generative models proliferate and concerns about training data provenance intensify, MIAs that work under realistic access constraints become increasingly critical. The unseen-class problem setting captures important properties
of real-world privacy auditing that have not been captured by
previous MIA threat models.
This setting is challenging, but our evaluation shows that quantile regression is a promising approach due to its generalization properties.
We encourage the community to develop
further MIAs that address the real constraints of privacy auditing in highly sensitive settings.
\section{Background and Preliminaries}
\vspace{-.05in}
\label{background}

We first formalize the membership inference attack (MIA) setting and introduce relevant attack methods.

We begin with the supervised learning setup. Let $\mathcal{D} \in \Delta(\mathcal{X} \times \mathcal{Y})$ denote the data distribution over input features $\mathcal{X}$ and labels $\mathcal{Y}$. The target model $f$ is trained on a dataset $D_{\text{priv}} \sim \mathcal{D}$ consisting of $n_{\text{priv}}$ labeled examples $(x_i, y_i)$. In the classification setting, we assume $\mathcal{Y}$ is a finite label set with $|\mathcal{Y}| = c$. The model $f$ outputs a vector of logits, i.e., $f : \mathcal{X} \to \mathbb{R}^c$.

In a membership inference attack, the adversary aims to determine whether a given \emph{target} example $(x, y)$ was part of the private training dataset $D_{\text{priv}}$. The adversary is typically assumed to have access to an auxiliary dataset $D_{\text{pub}} \sim \mathcal{D}$, which is disjoint from the private dataset, i.e., $D_{\text{pub}} \cap D_{\text{priv}} = \emptyset$. However, in this work we consider a practical setting where $D_{\text{pub}}$ is drawn from a more restricted sub-population.

\vspace{-.05in}
\paragraph{Setting: MIA with limited access to classes.} 
In practice, the adversary may only have access to samples from a 
\emph{subset} of the classes used to train the target model.
This can happen for a number of reasons.
For example, the background data may be drawn from a public source such as data available on the Internet, 
while the target model may include samples from private or proprietary data sources.
For auditors who want to run MIA to audit a model for potentially harmful content (as in the CSAM example described above~\citep{thiel2023identifying}),
that content may not be legally available to the auditor at large enough scale to train the attack.
Alternatively, the adversary may simply be resource-limited and unable to collect representative samples covering the space of data used to train the target model.

Let $Y_d \subseteq \mathcal{Y}$ denote the set of \emph{unseen} classes. In this setting, the adversary has access to a public dataset $D_{\text{pub}}'$ drawn from the conditional distribution $\mathcal{D}_{\neg Y_d} := \mathcal{D} \mid y \notin Y_d$, which contains only \emph{seen} classes. Despite this restriction, we aim to evaluate the performance of the membership inference attack (MIA) on target examples drawn from the full distribution $\mathcal{D}$. This setup allows us to study whether MIA methods trained on a restricted subset of classes can generalize to previously unseen classes from the original distribution.

We now introduce three classes of MIA methods. Each method relies on a \emph{score function} $s$ that assigns a numeric score to a target example $(x, y)$, intended to reflect the likelihood that $(x, y)$ was included in the training set for the model $f$. For example, in \citet{bertran2023scalable, lira}, an example of such a score function is based on logit differences:
\begin{equation}
s(x, y, f) = f(x)_y - \max_{y' \neq y} f(x)_{y'}
\end{equation}
\vspace{-.15in}

\vspace{-.05in}
\paragraph{1) Marginal baseline attack with a single threshold.}
The marginal baseline attack\citep{yeom2018privacy} (which we refer to as ``LOSS'' throughout the paper in keeping with \citep{lira}) is a simple yet widely used baseline for membership inference. Here, the attacker chooses a single threshold $\tau$ such that any example $(x, y, f)$ with $s(x, y) > \tau$ is predicted to be a member of $D_{\text{priv}}$, and otherwise it is predicted to be a non-member. In our setting, we compute this threshold based on the held-out public dataset $D'_{\text{pub}}$.
Specifically, the threshold $\tau$ is chosen to control the false positive rate (FPR) over $D'_{\text{pub}}$. Note that the FPR computed over $D'_{\text{pub}}$ may not accurately reflect the FPR under the distribution $\mathcal{D}$ due to the distribution shift.


\vspace{-.05in}
\paragraph{2) Shadow model-based attacks.} 

Unlike the simple marginal baseline attack, shadow model attacks consider performing MIA with per-example thresholds. In particular, each shadow model ~\citep{shokri2017membership, lira} constructs a reference distribution over model outputs to evaluate membership of a target example. Formally, the adversary trains $k$ shadow models $g_1, \ldots, g_k$, each solving the same classification task as the target model $f$, with architecture and training procedure identical to that of $f$, and trained using the same algorithm as $f$. 
Most SoTA attacks use such shadow models as a core component of the attack~\citep{lira, choquette2021label, zarifzadeh2023rmia}. 

In our setting, each shadow model is trained on random subsets drawn independently from the public dataset $D'_{pub}$. 
In keeping with the computationally tractable ``offline'' attack in ~\citet{lira}, for each target point,
we only use the set of shadow models for which the training
point is a \emph{nonmember}.~\footnote{~\citet{lira} evaluate both the online and offline
attacks and show that the difference in the ROC curves is minimal.}
The collection of shadow models allows the attacker to learn the conditional distribution of the scores:
\begin{align*}
    P_{\text{out}}: s(x, y, g) &\mid (x, y) \mbox{ is not used in training } g
\end{align*}
Given the target model $f$, the attacker decides membership based on the probability of the score $s(x, y, f)$  under $P_\mathrm{out}$. 

\vspace{-.05in}
\paragraph{3) Quantile regression attack.}  The quantile regression attack~\citep{bertran2023scalable, tang2023membership} offers an efficient alternative to shadow model approaches for membership inference. Rather than fitting a distribution over shadow model outputs, the attacker directly learns a function that maps input examples to score thresholds—thereby enabling {per-example} thresholds. Given a target FPR $\alpha$, the attacker trains a model $q_\alpha: \mathcal{X} \times \mathcal{Y} \rightarrow \mathbb{R}$ to estimate the $(1 - \alpha)$-quantile of the score distribution, conditioned on the input $(x, y)$.
 The model $q_\alpha$ is trained via minimizing the \emph{pinball loss} over a function class $\mathcal{H}$ on the public dataset
    \begin{equation}
    q_\alpha \in \arg\min_{q' \in \mathcal{H}} \, \mathbb{E}_{(x, y) \sim D'_{\text{pub}}} \left[ \text{PB}_{1 - \alpha}(q'(x), s(x, y, f))\right]
    \end{equation}
where $\text{PB}_{1-\alpha}$ is defined as $\text{PB}_{1-\alpha}(\hat{s}, s) = \max\{\alpha(\hat{s} - s), (1 - \alpha)(s - \hat{s})\}$.
This loss function is well known to elicit quantiles, in the same way that squared loss elicits means. Then the MIA predicts membership if $s(x, y, f) > q_\alpha(x)$.

Notably, this approach requires training only a single, lightweight model unlike shadow model attacks, which often demand replicating the full training pipeline and architecture of the target model.

\paragraph{Evaluation metrics.} Across our main results, we use \emph{true positive rate at low false-positive rate} as our main metric, 
where low false positive rates are 1\% and 0.1\%. 
This is in keeping with best practices recommended by~\cite{lira}.
\section{Shadow Models Fail on Unseen Classes}
\vspace{-.1in}
\label{sec:expts-shadow}

We first show that attacks that use shadow models fail
in the unseen-class setting.
Shadow models underpin the most popular and rigorous membership inference attacks, 
from the first proposals for membership inference~\citep{shokri2017membership} up to the current state of the art~\citep{zarifzadeh2023rmia}.

\vspace{-0.05in}
\paragraph{Setup.} We assume a setting where the target model 
is trained on half the training data,
the adversary has access to all but one class (from the remaining training data) to train the attack,
and the query points are drawn from the unseen class.

For each attack, we train 16 shadow models
using the ``offline'' variant of the LiRA algorithm
with a fixed, global variance estimate (following ~\citet{lira}).
We use the LiRA attack unmodified except that the shadow models'
training data excludes examples from the unseen class.

\begin{figure}[t!]
\centering
\begin{subfigure}{0.4\textwidth}
\includegraphics[width=\textwidth]{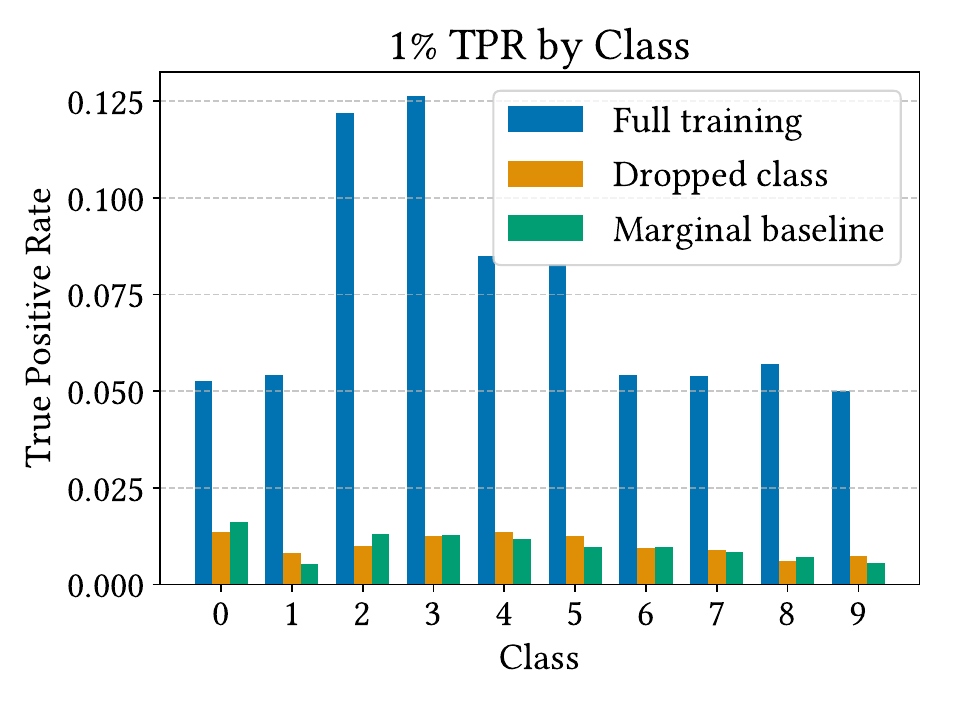}
\vspace{-.26in}
\caption{Results for CINIC-10.}
\end{subfigure}
\begin{subfigure}{0.4\textwidth}
\includegraphics[width=\textwidth]{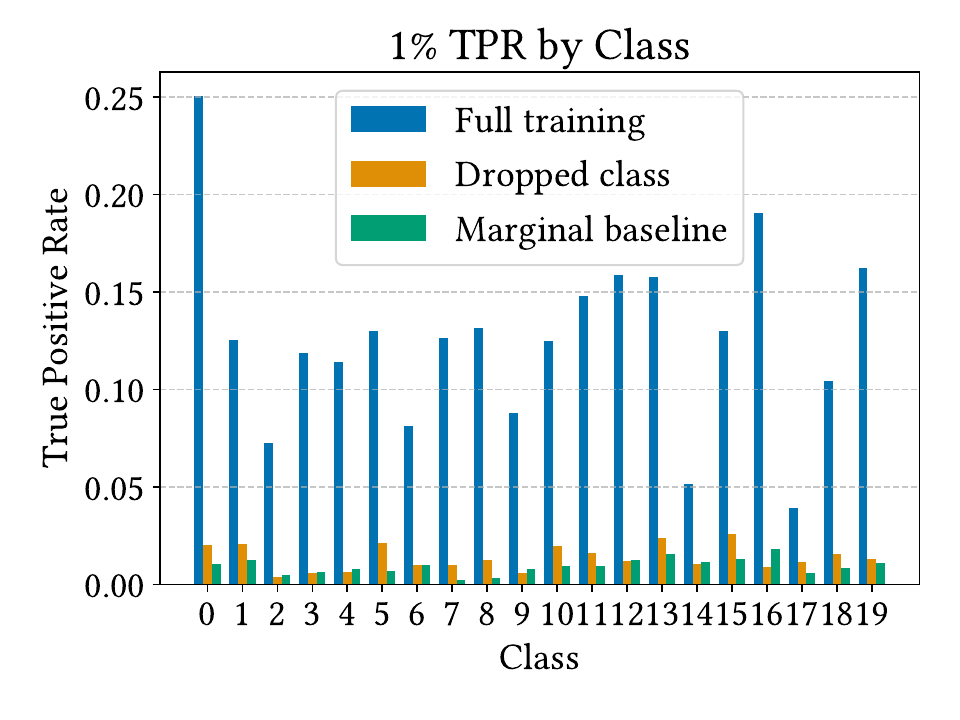}
\vspace{-.26in}
\caption{Results for CIFAR-100 (coarse labels).
}
\end{subfigure}
\setlength{\belowcaptionskip}{-0.3in}
\caption{True positive rates for shadow model attacks in the 1\% false positive rate regime for CINIC-10 and CIFAR-100 (we defer the 0.1\% regime to Appendix \ref{appx:shadowmodel-results}). Each bar represents the TPR on the indicated class. 
``Full training'' refers to the TPR on class $i$ when no classes are excluded from shadow model training.
In yellow, we plot the TPR when that class is excluded from shadow model training. The attack success degrades significantly under class exclusion, often performing worse than the marginal baseline (global threshold).}
\label{fig:shadow-degradation}
\end{figure}

We train shadow models on CINIC-10 and CIFAR-100 (using the 
superclass label set consisting of 20 classes).
In Figure~\ref{fig:shadow-degradation} we report the performance of shadow models
on queries from each class before and after dropping a single class at a time. We find that for both datasets, shadow models perform on par with a much weaker marginal baseline that does not learn per-example thresholds but rather fits a single threshold across all classes.

Intuitively, a model that does not see a class at training time will assign zero probability to that class label.
As a result, the shadow models' confidences for the true label on the missing class will be zero, 
resulting in many false positives at test time---any higher confidence reported by the target model would be considered significant.
This applies to both the LiRA attack as well as other
more recent attacks (such as RMIA~\citep{zarifzadeh2023rmia}) 
that judge membership using the score of the query on the target model
\emph{relative to} the score on reference models that have never seen the query class.
In the next section, we demonstrate a similar failure for RMIA.

For completeness, in Appendix~\ref{appx:shadowmodel-results} we also include shadow model results using the
difference between the top two logits (rather than the correct label)
as the score function.
We find that this score function reduces the baseline performance of the attack.
Thus, for the remainder of our comparisons,
we use the true label confidence as the score function for shadow models.
\section{Quantile Regression Attacks \\ for Unseen Classes}
\label{sec:expts-qr}
\vspace{-.05in}
Shadow models are fundamentally constrained because they must solve the same learning problem as the target model
in order to estimate the score distribution of the target model on a given example.
Quantile regression attacks~\citep{bertran2023scalable}
were originally proposed as an MIA that can provide a provable guarantee on the attack's false-positive rate by predicting an $\alpha$-quantile of the score distribution.

In this section, we make a novel observation -- that quantile regression attacks also \emph{generalize} well to unseen classes,
because they learn features that distinguish members from nonmembers.
Unlike shadow model-based attacks, which judge membership by evaluating the score of the target model relative to the score of a reference model,
quantile regression attacks directly predict thresholds from target model scores on the background distribution.
This gives them an unexpected advantage in the unseen-class setting.

Our results give a new perspective on the strengths and weaknesses of shadow models and indicate that attacks that learn membership predictors,
including but not limited to quantile regression attacks,
may be much stronger candidates for \emph{real-world} membership inference settings where the data to query is sensitive and not available for training.

\subsection{Unseen classes for image datasets.}
\vspace{-.05in}

Our first set of results is on image classification models.
This setting models scenarios such as detecting child sexual abuse material (CSAM) in image models.

\begin{figure}[h!]
    \centering
    \begin{subfigure}[b]{0.4\textwidth}
        \centering
        \includegraphics[width=\textwidth]{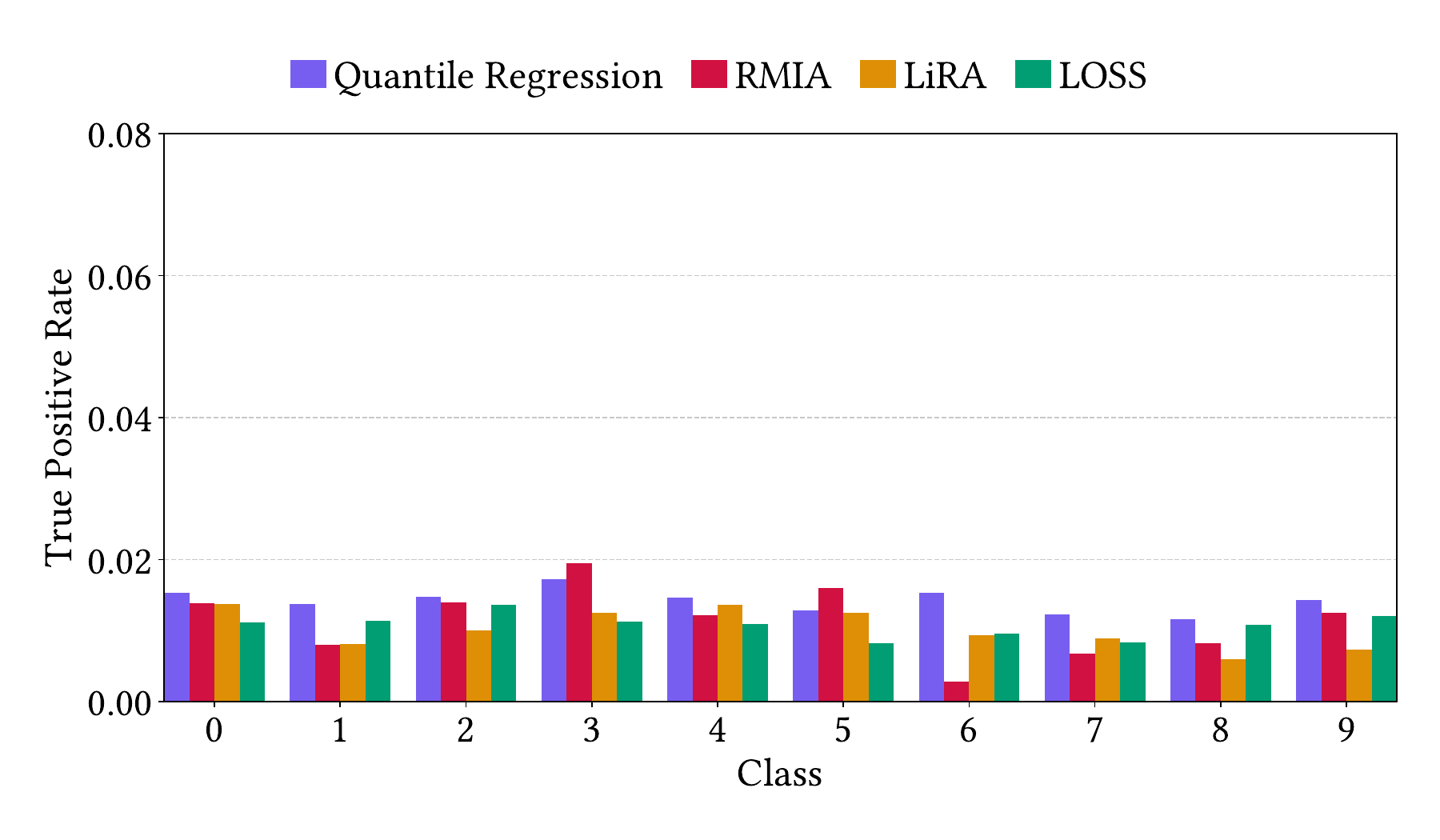} 
        \caption{TPR for the dropped out class, CINIC-10}
        \label{fig:qr_cifar10_fpr1}
    \end{subfigure}\hfill%
    \begin{subfigure}[b]{0.4\textwidth}
        \centering
        \includegraphics[width=\textwidth]{figures/qr_cifar100_fpr1.pdf} 
        \caption{TPR for the dropped out superclass, CIFAR-100}
        \label{fig:qr_cifar100_fpr1}
    \end{subfigure}
    \setlength{\belowcaptionskip}{-0.15in}
    \caption{True positive rates in the low false positive regime for CINIC-10 and CIFAR-100 (superclass set) on each unseen class. 
    Each bar represents the true positive rate on class $i$ when class $i$ is dropped from the attack training set. We only report results at 1\% FPR; the results at 0.1\% FPR are not meaningful due to the small sample size of the validation set on a single class (1000 samples).
    While quantile regression attacks have only a small advantage
    over shadow models on CINIC-10 (see Figure \ref{fig:gmm}), they achieve up to 11$\times$
    higher TPR than shadow models on CIFAR-100. Note that
    the yellow bars (LiRA) correspond to the respective yellow
    bars in Figure~\ref{fig:shadow-degradation}.}
    \label{fig:qr_cifar10}
\end{figure}

\vspace{-0.1in}
\paragraph{Setup.} We first study an identical setting where the target Resnet50 base model is trained on half the training data, the adversary has access to all but one class from the remaining training data to train the attack, and the query points are drawn from the unseen class.

As baselines, we use the LiRA attack described above, as well as the state of the art RMIA attack~\citep{zarifzadeh2023rmia}.~\footnote{
Notably, in addition to training reference models,
the RMIA attack also uses samples from the background distribution when \emph{evaluating} the score function at query time.
It is generally unrealistic to assume that the adversary has access to enough unseen-class, nonmember samples to perform this evaluation.
However, in order to give RMIA the \emph{most advantage}, 
we \emph{include} unseen class samples at evaluation time
(but not when training reference models, as is the case for shadow models and quantile regression).}

For each attack, we train a single quantile regression model on the remaining training data, excluding the unseen class and keeping the validation set as heldout public data for evaluating FPR. Pinball loss is notoriously difficult to minimize, so for training stability, we follow  \citet{bertran2023scalable} and train the network to fit a Gaussian (mean and variance) conditioned on each sample instead of directly using pinball loss to predict quantiles. 

Our final models for the CINIC-10 and CIFAR-100 attacks are ConvNext-Tiny-224 models trained for 30 epochs, with the Adam optimizer, batch size 16, and learning rate of 1e-4. We find that early stopping does not improve the attacks. To make our attack agnostic to the true label, we modify the attack from \citet{bertran2023scalable} and use the difference between the top two logits as our score function: 
\[
s(x,y,g) = \max f(x) - \max_{y' \neq \max f(x)} f(x)_{y'}.
\]
Using this score function improves quantile regression performance in the class dropout setting as the learned attack no longer requires knowledge of the true label. 

\vspace{-.05in}
\paragraph{Class dropout: CINIC-10 and CIFAR-100.} We find that for both CINIC-10 and CIFAR-100 (using the superclass label set), quantile regression strictly outperforms the marginal baseline and shadow models under class dropout at 1\% FPR. In this setting, we train the attack model on all classes except the unseen class. The TPR and FPR are evaluated on the held-out class. 

Results are similar on CIFAR-100 but even more pronounced. On average across all 20 superclasses, LiRA achieves only 1.4\% TPR at 1\% FPR on unseen data while quantile regression achieves 3.8\% TPR at 1\% FPR (a 2.7$\times$ improvement). In the unseen class setting, quantile regression takes less time to train and outperforms LiRA.
We find that RMIA often performs \emph{worse} than the LiRA attack. 
This is because RMIA's score function exaggerates the target model's confidence on the unseen class,
resulting in more false positives in this setting and very low TPRs in the low-FPR regime.

These trends also hold on ResNet-18 and Vision Transformer base models; detailed results are found in Appendix \ref{appx:image-architecture-results}. In addition, we also provide AUC results in Appendix~\ref{appx:auc-results} and full ROC curves in Appendix ~\ref{appx:roc}.


\vspace{-.05in}
\paragraph{Data scarcity: ImageNet.} Our results on CINIC-10 and CIFAR-100 are limited to 
a single unseen class at a time,
which models the impact of a minority subpopulation
(such as sensitive or harmful content) missing from the attack data.
Another realistic scenario where the auditor or adversary might not have access to subclasses is the \emph{data scarcity}
setting, where the model might be trained using a much larger and potentially proprietary dataset
while a resource-limited auditor only has a fraction of similar data available.

\begin{figure}[h!]
    \centering
    \begin{subfigure}[t]{0.32\textwidth}
        \centering
        \includegraphics[width=\textwidth]{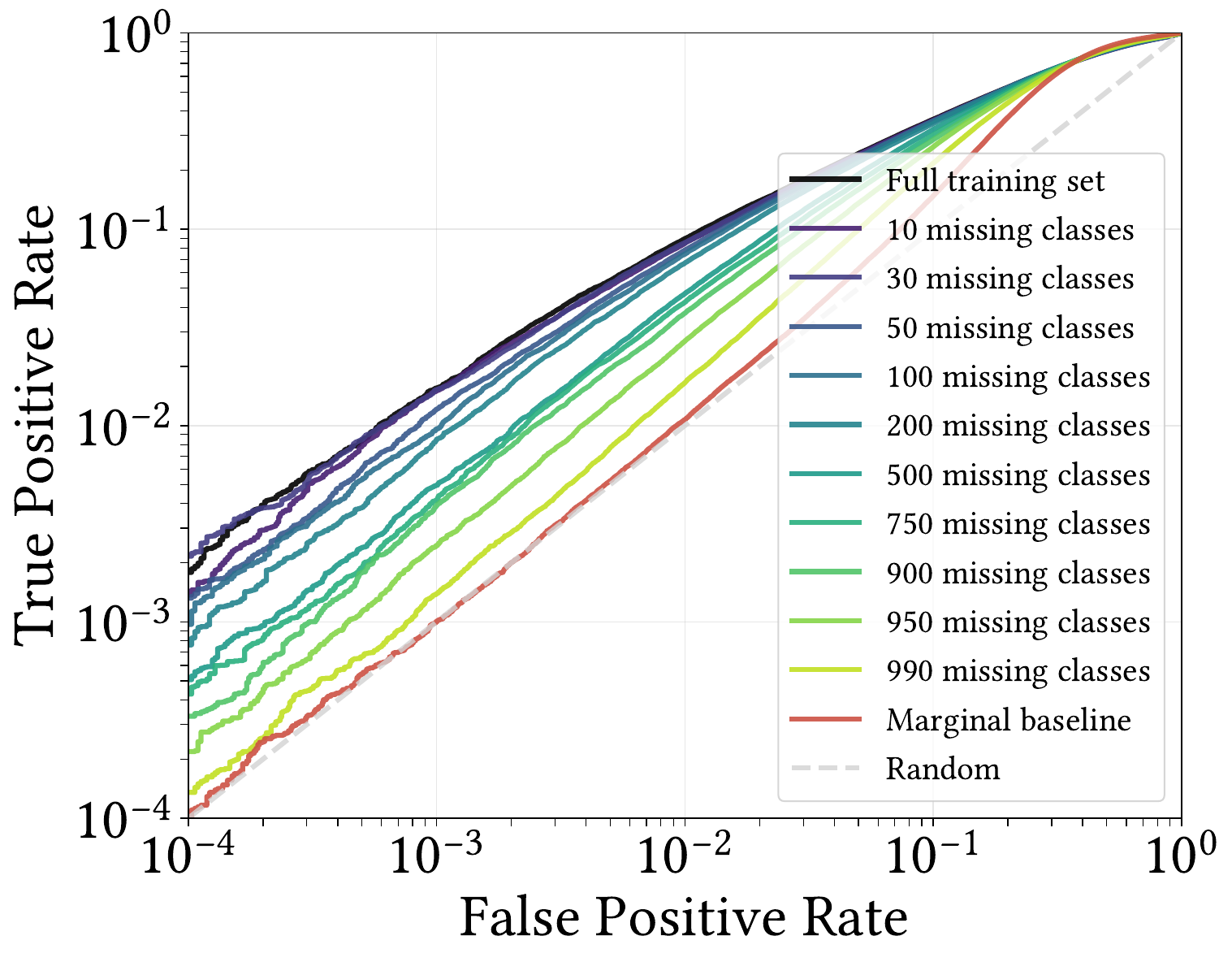}
        \caption{ROC curve for class drop experiment on ImageNet.}
        \label{fig:imagenet_classdrop_roc}
    \end{subfigure}
    \hfill
    \begin{subfigure}[t]{0.32\textwidth}
        \centering
        \includegraphics[width=\textwidth]{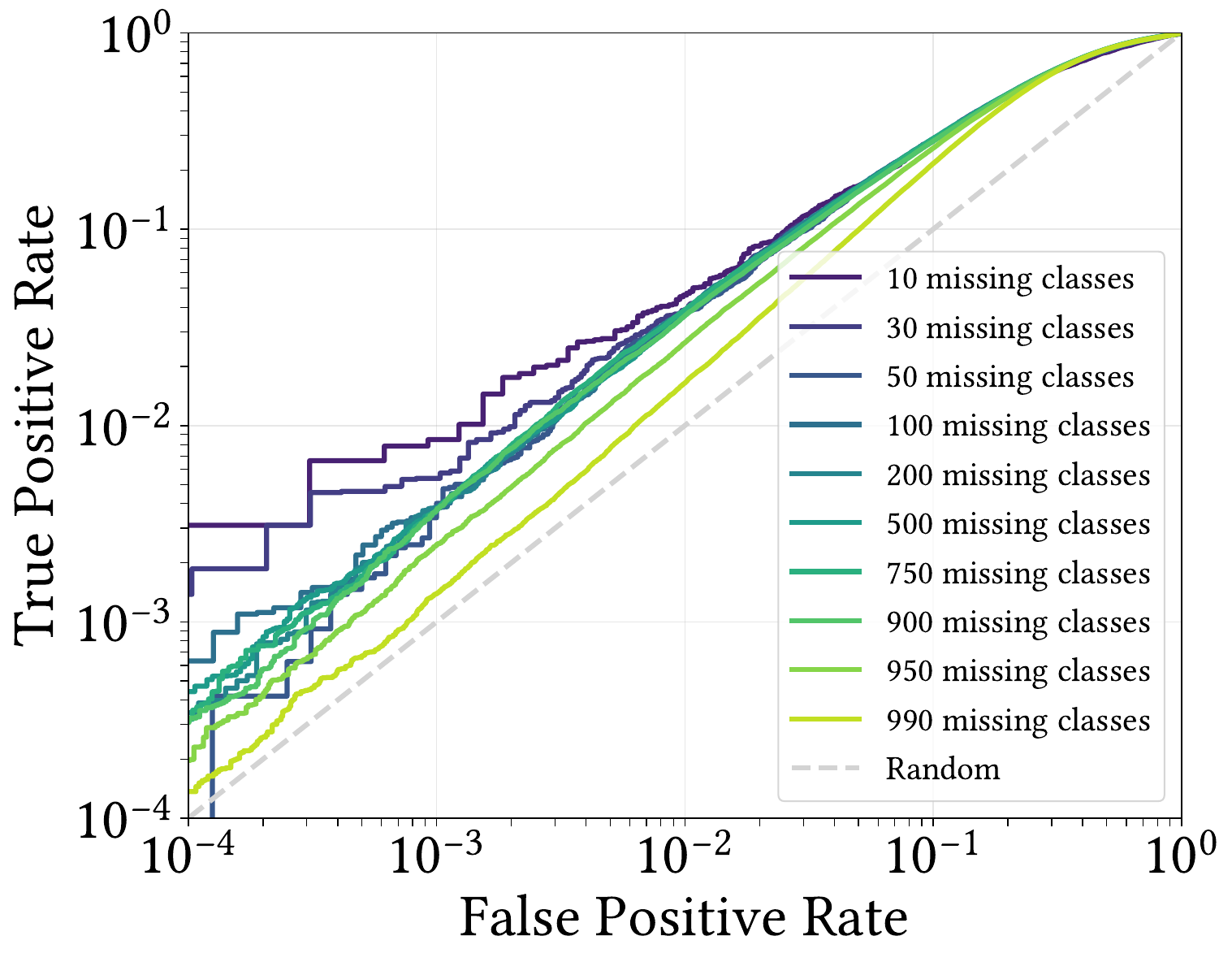}
        \caption{Unseen class ROC curve for class drop experiment on ImageNet.}
        \label{fig:imagenet_classdrop_roc_ood}
    \end{subfigure}
    \hfill
     \begin{subfigure}[t]{0.32\textwidth}
        \centering
        \includegraphics[width=\textwidth]{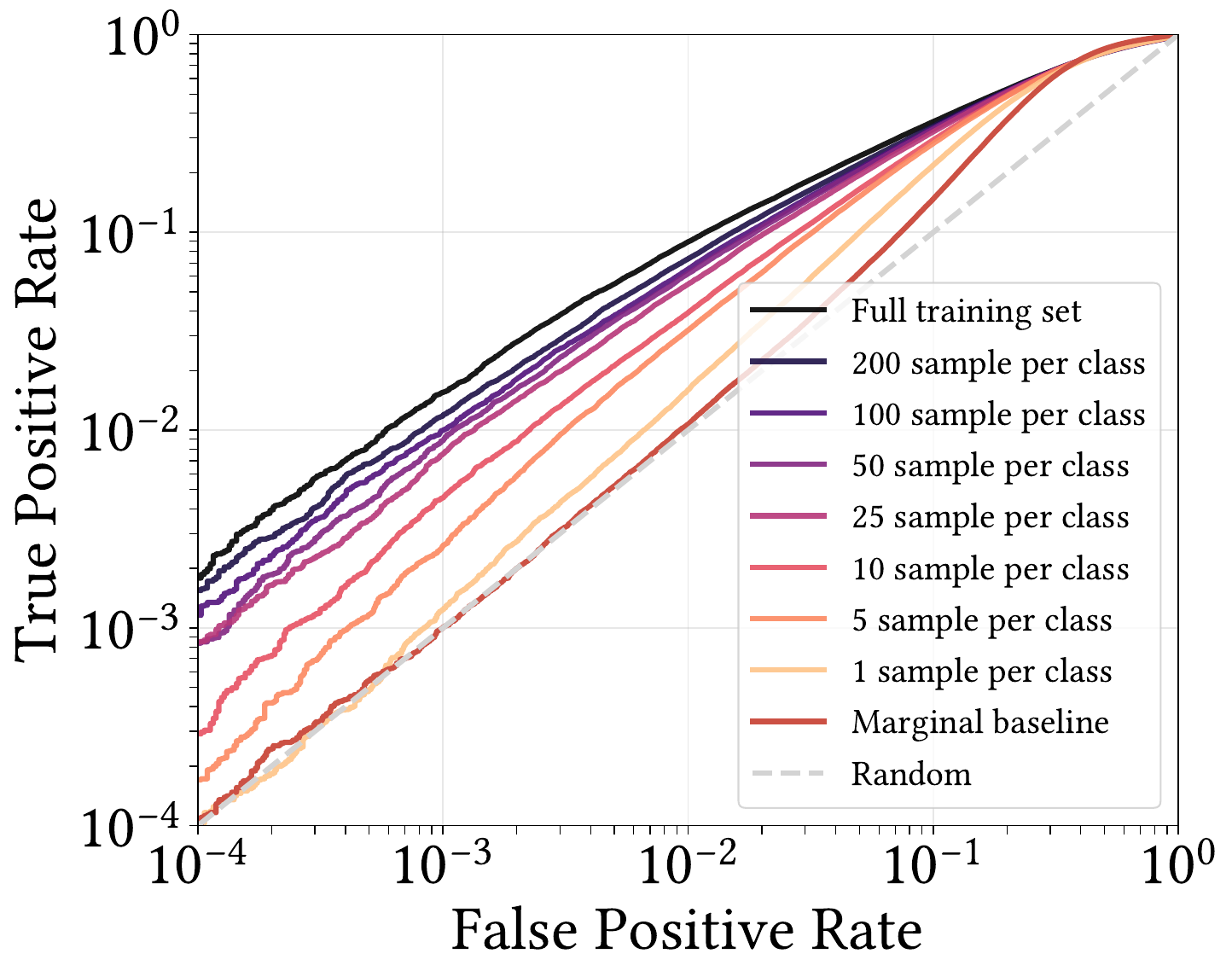}
        \caption{ROC curve for sample drop experiment on ImageNet.}
        \label{fig:imagenet_sampledrop_roc}
    \end{subfigure}
    \vspace{-.08in}
    \caption{ROC curves for class and sample drop experiments on ImageNet. Enlarged versions of the plots are provided in Appendix \ref{appx:enlarged-plots}.}
    \label{fig:imagenet_roc_combined}
    \vspace{-.1in}
\end{figure}

To simulate this, we attack a model trained on the much larger ImageNet dataset. We use the same quantile regression architecture and hyperparameters as described for CINIC-10 and CIFAR-100. 
In Figure \ref{fig:imagenet_classdrop_roc}, 
we show the ROC curves for ImageNet (on the full data distribution) with a sweep from 10 to 990 classes missing from the attack training set.
Perhaps surprisingly, the quantile regression attack outperforms the marginal baseline with as many as 990 out of 1000 classes left unseen. 

One might ask whether this effect is simply due to averaging over both missing and in-distribution classes. In Figure \ref{fig:imagenet_classdrop_roc_ood},  we plot attack performance for the missing classes alone. The performance on unseen classes remains fixed around 3.8\% TPR at 1\% FPR (0.4\% TPR at 0.1\% FPR) when anywhere from 100 to 900 classes are removed, and remains significantly above the marginal baseline even with 990 classes removed.

Finally, another realistic setting of data scarcity is one where 
examples from all classes are available, but
where the auditor only has very few samples from each class. 
We also evaluate ImageNet in this scarce-sample setting where only $k$ samples are retained from each class. Quantile regression outperforms the marginal baseline (Figure \ref{fig:imagenet_sampledrop_roc}) even when training data is severely limited, retaining 3.9\% TPR at 1\% FPR given as few as 10 samples per class (compared to 9.0\% TPR at 1\% when trained on all the data; on average, 401 samples per class).

We present similar results for data scarcity on CIFAR-10 and CIFAR-100 in  Appendix \ref{appx:multiclass-drop-results}. Quantile regression achieves, for example, ~3\% TPR at 1\% FPR even when half of the classes are dropped from CIFAR-100. 

\vspace{-.1in}
\subsection{Unseen classes for tabular and text datasets.}\vspace{-.05in}
The unseen class scenario also arises in tabular and text settings.
For example, consider auditing medical records in which subsets of records may be inaccessible to the auditor due to privacy, legal, or ethical restrictions~\citep{tramer2022position},
or population survey data in which some records might belong to sensitive minority groups and cannot be made public~\citep{steed2024quantifying}.

Next, we show that shadow models and global thresholding fail on unseen classes in tabular and text classification settings as well. For these experiments, we compare quantile regression to RMIA, the state-of-the-art shadow model approach from \citet{zarifzadeh2023rmia}. 

\begin{figure}[h!]
    \centering
    \begin{subfigure}[t]{0.4\textwidth}
        \centering
        \includegraphics[width=\textwidth]{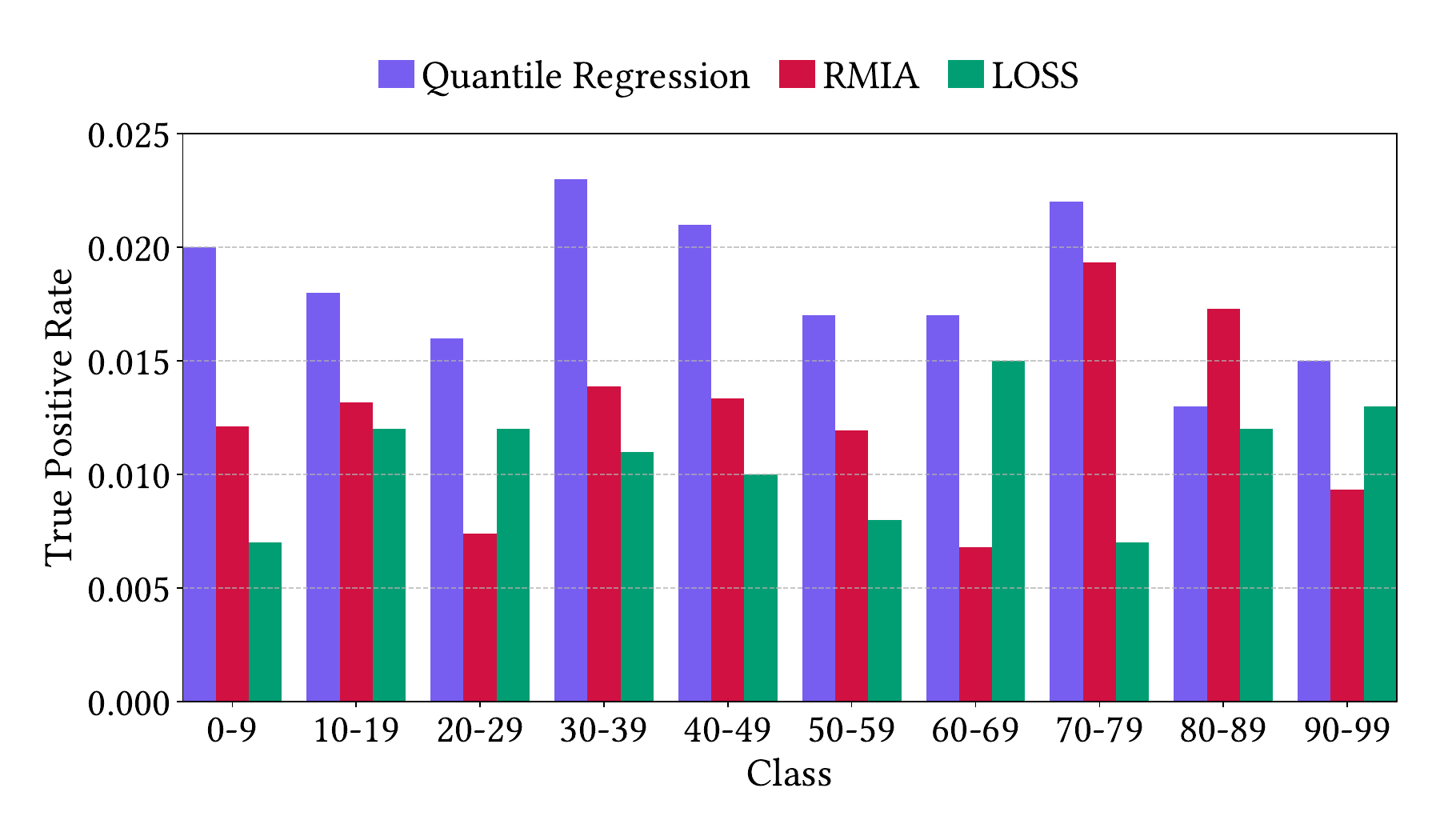} 
            \setlength{\abovecaptionskip}{-1em}
    \setlength{\belowcaptionskip}{-0.3em}
        \caption{TPR for the dropped out classes, Texas}
        \label{fig:qr_texas_fpr1}
    \end{subfigure}\hfill%
    \begin{subfigure}[t]{0.4\textwidth}
        \centering
        \includegraphics[width=\textwidth]{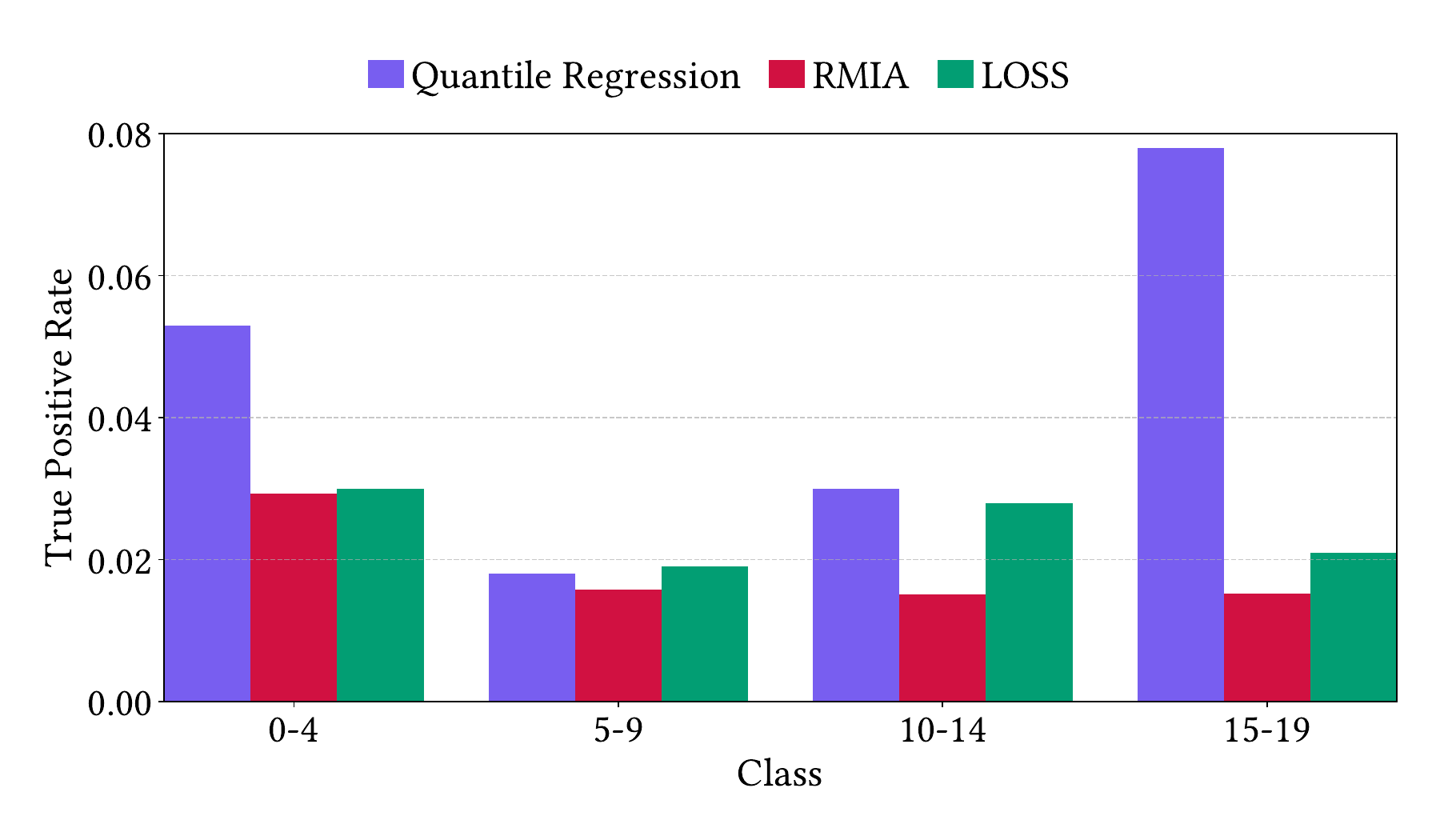} 
                    \setlength{\abovecaptionskip}{-1em}
            \setlength{\belowcaptionskip}{-0.3em}
        \caption{TPR for the dropped out class, 20 Newsgroups}
        \label{fig:qr_ng20_fpr1}
    \end{subfigure}
        \setlength{\belowcaptionskip}{-0.2in}
    \caption{True positive rates in the low false positive regime for Texas (tabular) and 20 Newsgroups (text) on sets of unseen classes. Each bar represents the true positive rate on classes $C$ when $C$ are dropped from the attack training set. We only report results at 1\% FPR; the results at 0.1\% FPR are not meaningful due to the small sample size of the validation set on a single class. Quantile regression attacks achieve up to 2$\times$
    higher TPR than shadow models on Texas and up to 6$\times$ higher TPR on 20 Newsgroups.}
    \label{fig:qr_text_tabular}
\end{figure}

\vspace{-.1in}
\paragraph{Setup.} For text classification, we fine‑tune GPT‑2~\citep{Radford2019Language} for 20-NewsGroups classification. 
Inputs are tokenized with truncation and padding to a fixed length of 256 tokens. We train with AdamW, batch size 32, learning rate 1e‑3, weight decay 5e-4, 500 warmup steps, for 30 epochs. For Texas tabular classification, we follow \cite{zarifzadeh2023rmia} and train a 2-layer MLP base model. For text and tabular, our attack models are also 2-layer MLPs (hidden sizes 256, 128). For text, the input is an embedding bag over the 256 input tokens. The hyperparameters and score function match the image auditing attacks described above.

\vspace{-.1in}
\paragraph{Class dropout: Texas and 20-NewsGroups.} We find that on tabular and text data, quantile regression again strictly outperforms the marginal baseline and RMIA shadow models under class dropout at 1\% FPR. The same is true for AUC, and full results can be found in Appendix~\ref{appx:auc-results}. We drop 10\% of classes for both datasets (10/100 for Texas, 2/20 for 20-NewsGroups). As described above, the TPR and FPR are evaluated on the held-out class.
\vspace{-0.1in}
\subsection{Defenses}

In general, models that one would wish to audit are \emph{not} deployed with typical MIA defenses such as differential privacy ~\citep{thornsafetybydesign}.
Nevertheless, for completeness, we evaluate our method 
in the unseen-class setting on models trained with one of two
defenses: weight decay (equivalent to $\ell_2$ regularization for SGD) and differential privacy.
We note that differential privacy has been noted by several previous works to be a strong (provable) MIA defense that prevents most MIA methods from succeeding, but is highly impractical~\citep{leino2020stolen, choquette2021label, lira, li2021membership} due to the difficulty of achieving reasonable test accuracy on models trained from scratch.

We provide these results in Appendix~\ref{appx:defenses}. Predictably,
we find that weight decay is less effective than DP at preventing MIA from succeeding and more regularization leads to worse MIA success.
In reasonable defense regimes where the TPR at low FPR exceeds random guessing,
quantile regression still outperforms the baseline 
in the unseen class setting.
\section{Theoretical Model}
\label{sec:theory}
\vspace{-.05in}
Our empirical results show that the quantile regression attack can achieve nontrivial 
accuracy even under extreme data scarcity when training the attack model.
However, it is not clear from our empirical results \emph{when} we might expect quantile regression to succeed.

As a step toward understanding our results, 
we prove a ``transferability'' theorem for quantile regressors.
Intuitively, this theorem states that if the distribution of sample embeddings under the quantile regression model
with and without the unseen classes is ``similar,''
then the FPR guarantee of the quantile regressor trained on only the seen classes should also hold on the full query set.

\begin{definition}[Pinball loss]
For a quantile level $\alpha \in (0,1)$, the \emph{pinball loss} (also known as the check loss) for a prediction $\hat{s}$ and true outcome $s$ is defined as:
\[
\ell_\alpha(\hat{s}, s) 
= (\alpha - \mathbf{1}\{s < \hat{s}\})(s - \hat{s}) = 
\begin{cases}
\alpha (s - \hat{s}) & \text{if } s \ge \hat{s}, \\
(1 - \alpha)(\hat{s} - s) & \text{if } s < \hat{s}.
\end{cases}
\]
\end{definition}

\vspace{-.1in}
\paragraph{Quantile Regression Predictor.}  
We consider a class of linear quantile regression predictors:
\[
q_\alpha(x) = \langle \phi(x), w \rangle,
\]
where $\phi: \mathcal{X} \to \mathbb{R}^d$ is a fixed feature mapping, and $w \in \mathcal{W}$ is a weight vector. For any distribution $P$ over $\mathcal{X} \times \mathcal{S}$, we will write $P_\phi$ to denote its induced distribution over $(\phi(x), s)$.

We will focus on the case where $\mathcal{W} = \mathbb{R}^d$, but should generalize to more constrained set of weights later. We adapt the multi-accuracy definition \citep{roth2022uncertain, hebert2018multicalibration} to our specific setting with a feature mapping.

\begin{definition}[Multi-Accuracy for Quantile Prediction]
A predictor $q_\alpha: \mathcal{X} \to \mathbb{R}$ is said to be \emph{$(\mathcal{W}, \phi, \varepsilon)$-multi-accurate} for quantile level $\alpha$ with respect to distribution $P$ if, for every $w \in \mathcal{W}$,
\[
\left| \mathbb{E}_{(x,s) \sim P} \left[ \langle w, \phi(x)\rangle \cdot \left( \mathbf{1}\{s < q_\alpha(x)\} - \alpha \right) \right] \right| \leq \varepsilon.
\]
\end{definition}

We now show that multi-accuracy, when instantiated for quantile prediction, provides a sufficient condition for calibration to transfer across distributions. We consider the setting where a quantile predictor is trained on distribution \( P \) and deployed on a shifted distribution \( Q \). The theorem below shows that if the feature representation captures the density ratio between \( P \) and \( Q \) via a linear function, then the learned predictor remains calibrated at the target quantile level under \( Q \). This result can be viewed as a specialized instance of the \emph{universal adaptability} framework of \citet{kim2022universal}, tailored to multi-accurate quantile predictors derived from empirical risk minimization.

\begin{theorem}[Transferability of Quantile Predictors]
\label{thm:transferability}
Let $P$ and $Q$ be distributions over $(x, s)$, and let $\phi: \mathcal{X} \to \mathbb{R}^d$ be a fixed feature map. Suppose we learn a linear quantile predictor $q_\alpha(x) = \langle \phi(x), w^* \rangle$ by minimizing the expected pinball loss under $P$:
\[
w^* \in \arg\min_{w \in \mathcal{W}} \mathbb{E}_{(x, s) \sim P}[\ell_\alpha(\langle \phi(x), w \rangle, s)].
\]
Assume that the density ratio between $Q$ and $P$ satisfies:
\begin{align*}
\frac{dQ_\phi(\phi(x), s)}{dP_\phi(\phi(x), s)} = \langle \phi(x), v \rangle \quad \text{for some } v \in \mathcal{W}, \\
\text{ and for all } (\phi(x), s) \in \operatorname{supp}(Q_\phi).
\end{align*}
Then the learned predictor $q_\alpha$ is calibrated under distribution $Q$ at quantile level $\alpha$:
\[
\mathbb{E}_{(x, s) \sim Q} \left[ \mathbf{1}\{s < q_\alpha(x)\} - \alpha \right] = 0.
\]
\end{theorem}

We defer the proof to Appendix~\ref{appx:proof}.

Intuitively, the transferability theorem states that
when there exists a linear transformation between the feature representation of the (unseen-class) training distribution and the feature representation of the full distribution,
the false-positive-rate guarantee of the quantile regressor
over the training distribution
will also hold for the test (full) distribution.
The assumption of optimality in the last layer is mild, since the pinball loss is convex in \( w \) given fixed features.


\vspace{-0.1in}
\subsection{Empirical Estimates}
\vspace{-.1in}

\begin{figure}[t!]
    \centering
    
    \begin{subfigure}[t]{0.32\textwidth}
        \centering
        \includegraphics[width=\textwidth]{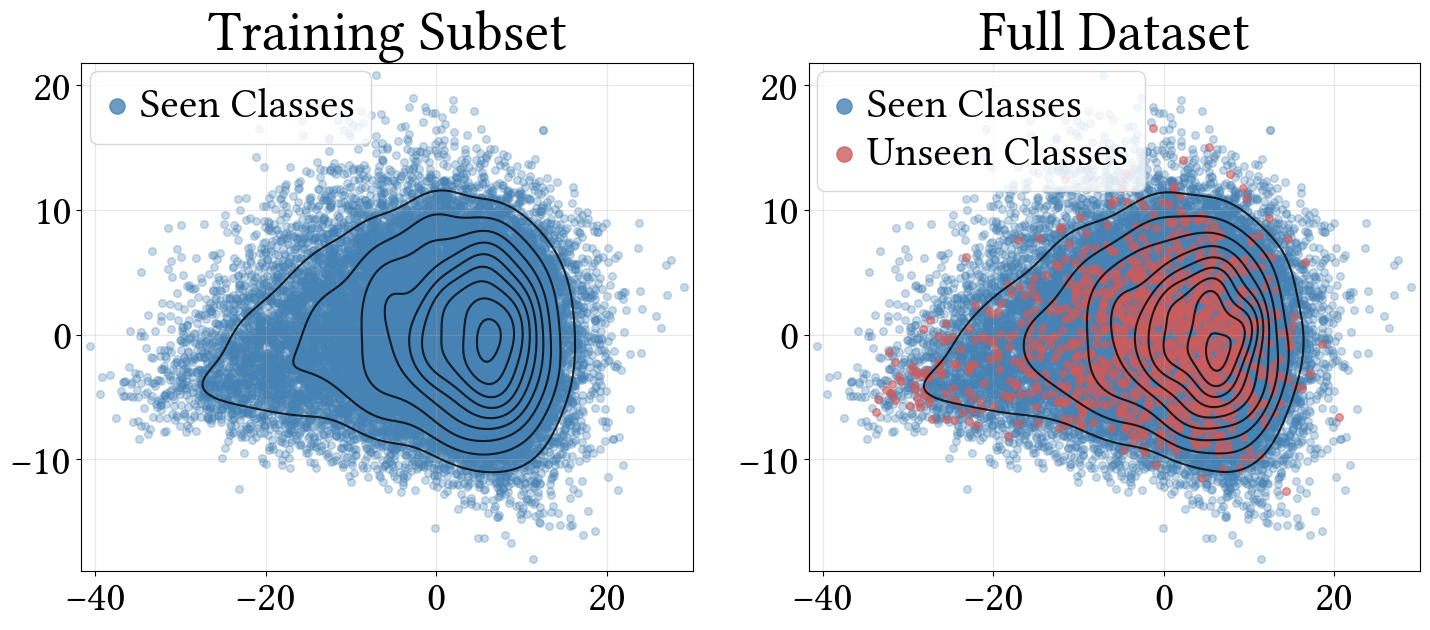}
        \caption{Gaussian mixtures fit to CIFAR-100 data with one unseen class.}
        \label{fig:gmm_c20}
    \end{subfigure}
    \hfill
    \begin{subfigure}[t]{0.32\textwidth}
        \centering
        \includegraphics[width=\textwidth]{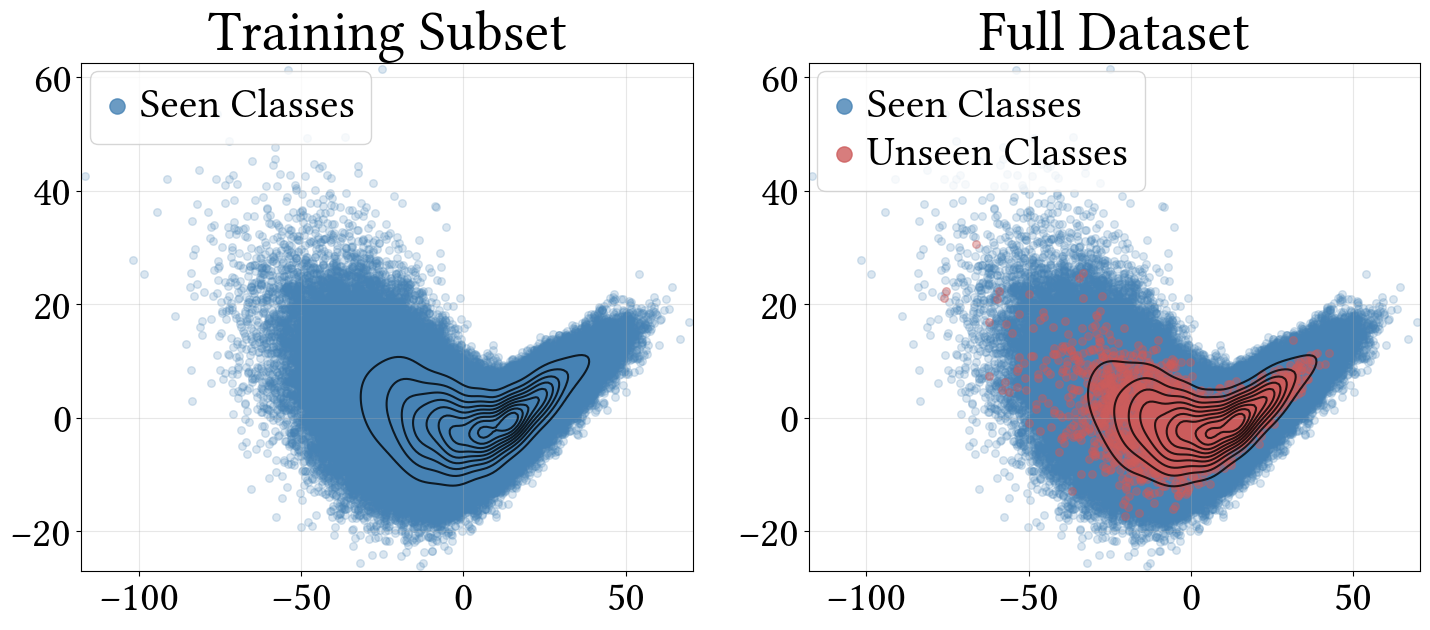}
        \caption{Gaussian mixtures fit to Imagenet data with ten unseen classes.}
        \label{fig:gmm_imagenet}
    \end{subfigure}\hfill
    \begin{subfigure}[t]{0.32\textwidth}
        \centering
        \includegraphics[width=\textwidth]{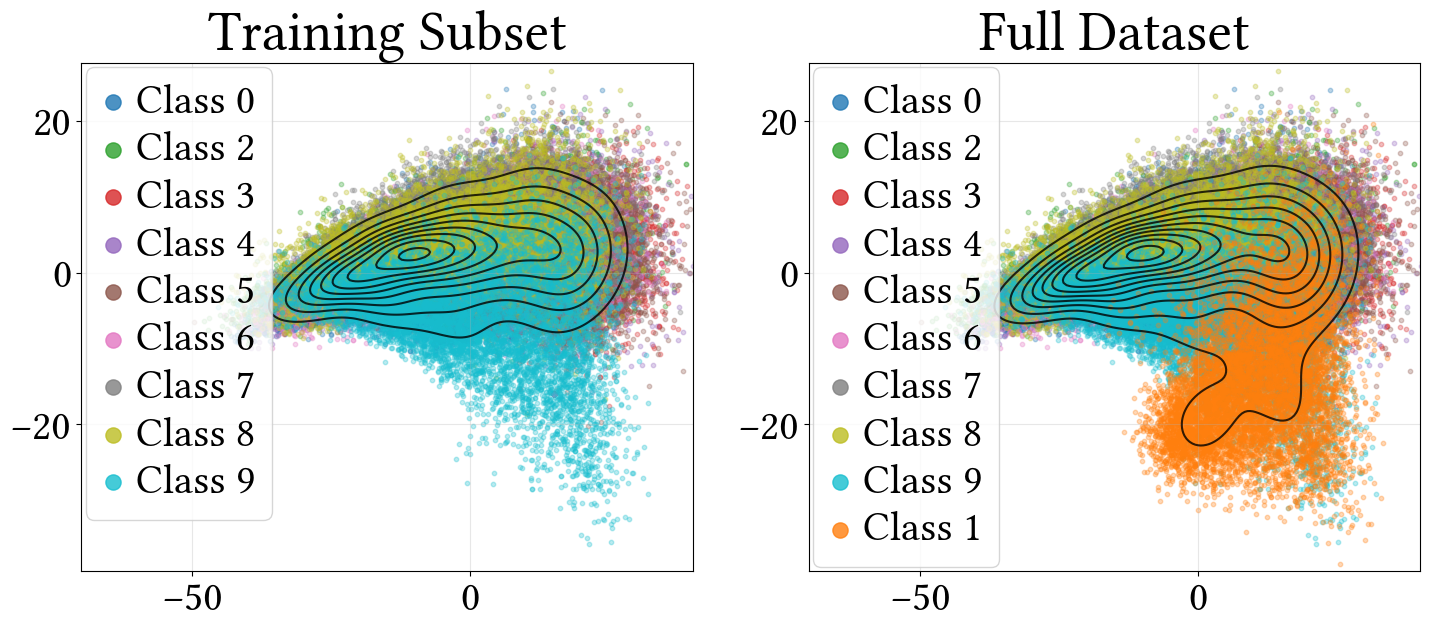}
        \caption{Gaussian mixtures fit to CINIC-10 data with one unseen class.}
        \label{fig:gmm_c10}
    \end{subfigure}
    \setlength{\belowcaptionskip}{-2em}
    \caption{Visualization of Gaussian mixture models fit to 
    the (dimension-reduced) embeddings learned by the quantile regression models trained on
    on subsets of CINIC-10, CIFAR-100, and Imagenet.
    Dropping a class largely does not change the distribution over embeddings for CIFAR-100 and Imagenet,
    where we observe that quantile regression is the most effective.
    }
    \label{fig:gmm}
\end{figure}

To understand the implications of this theorem in our setting,
we can treat the quantile regressor as a feature extractor
using the final layer before the prediction step 
to generate embeddings $\phi(x)$ for each sample in the dataset.
We compare the distribution over last-layer embeddings in the
attack training set (the seen subset of classes)
with the distribution over last-layer embeddings in the evaluation set.

Thus, in order to show that the theorem statement holds,
we need to measure the density ratio between the
attack training distribution and the test distribution.
In practice, we approximate the density ratio using Gaussian mixture models 
trained over a low-dimensional projection of the embeddings
(the top two PCA components) rather than the original embeddings
due to the relatively low sample size. These Gaussians are visualized in Figure \ref{fig:gmm}.

Using these approximations, we explicitly measure the density ratio over the test set and fit a linear layer
on top of the representation learned by the quantile regressor.
We find that the linear fit improves as the dataset size and diversity increases: 
the MSE of the linear model on CINIC-10 is 8.11e-3, 
on CIFAR-100 is 2.10e-3, and on Imagenet 1.85e-3. This experiment provides some validation of the empirical results
we see in practice and may expect based on the visual representation of the image embeddings:
the results of quantile regression are weakest on CINIC-10
(where the linear fit is comparatively worse)
and best on ImageNet (where the linear fit is best).
\section{Related Work}
\label{related_Work}

\paragraph{Membership inference attacks.}  
Shadow models were initially proposed by~\citet{shokri2017membership}
and refined by~\cite{lira}. 
Membership inference attacks based on quantile regression were introduced by~\citet{bertran2023scalable} 
for classification tasks.
Follow-up work~\citep{tang2023membership, carlini2023extracting}
extended this approach to the diffusion model setting
using a score metric inspired by~\citet{secmi}. Others~\citep{zhang2024membership, hayes2025strong}
have made progress towards applying these methods to large language models.

A number of earlier works propose using a single global score threshold
across examples rather than fitting per-example thresholds~\citep{shokri2017membership, yeom2018privacy, song2019privacy, secmi}.
Later work~\citep{watson2021importance, lira} has shown that these global score thresholds perform poorly compared to thresholds calibrated to example difficulty,
particularly in the low-false positive rate regime.

\paragraph{MIA under distribution shift.}
A number of prior works have studied the effectiveness
of shadow model attacks under varying distribution shift, 
although none of these study the unseen class setting g we study in our work.
~\citet{yichuan2024assessing} study a setting in which the attacker has access to data encompassing a \emph{superset} of the target labels,
as well as attribute shift corresponding to subpopulation reweighting.
~\citet{liu2022membership} study distribution shift between CIFAR-10 and CINIC-10,
which have the same label set.
Related attacks exist that do not use shadow models,
but do not apply in our setting:
~\cite{nasr2019comprehensive} is a white-box attack, and
~\cite{mattern2023membership, shi2023detecting} 
are attacks on generative text models.
~\cite{choquette2021label} requires shadow models.
Earlier works ~\citep{lira, shokri2017membership, ye2022enhanced}
have empirically studied how shadow models perform when the target architecture or training procedure are not known exactly,
but do not study the unseen-class setting.



\section*{Impact Statement}

Our work is directly motivated by addressing an important problem in ethical and safe ML. We worked with practitioners in the field to formalize the constraints they face in auditing models for sensitive data, particularly CSAM. The broader impact of our work includes (1) formalizing this problem and disseminating it to researchers to improve upon the methods proposed, (2) formalizing and evaluating why existing methods considered state of the art cannot perform well under these constraints, and (3) demonstrating an alternative method that has properties allowing it to perform well under these constraints. Our hope is that our work will be a stepping stone towards realistic and practical methods that can be used by law enforcement, model providers, and similar organizations to ensure that models are not trained on harmful and illegal content.
\section*{Acknowledgements}

We are extremely grateful to Rebecca Portnoff for  insightful discussions regarding realistic data access models for CSAM detection.
Thanks also to Shuai Tang, Martin Bertran, and Matteo Boglioni for discussions on quantile regression attacks and Alex Robey for helpful feedback on presentation.

This work was supported in part by the National Science Foundation grants IIS2145670 and CCF2107024, and funding from Amazon, Apple, Google, Intel, Meta, and the CyLab Security and Privacy Institute. Any opinions, findings and conclusions or recommendations expressed in this material are those of the author(s) and do not necessarily reflect the views of any of these funding agencies. P.T. was supported by a Carnegie Bosch Institute fellowship.  Z.S.W. was in part supported by NSF Awards \#1763786 and \#2339775.

\bibliography{miashift}

@inproceedings{lira,
  title={Membership inference attacks from first principles},
  author={Carlini, Nicholas and Chien, Steve and Nasr, Milad and Song, Shuang and Terzis, Andreas and Tramer, Florian},
  booktitle={2022 IEEE symposium on security and privacy (SP)},
  pages={1897--1914},
  year={2022},
  organization={IEEE}
}

@inproceedings{hebert2018multicalibration,
  title={Multicalibration: Calibration for the (Computationally-Identifiable) Masses},
  author={Hebert-Johnson, Ursula and Kim, Michael P. and Reingold, Omer and Rothblum, Guy N.},
  booktitle={Proceedings of the 35th International Conference on Machine Learning (ICML)},
  pages={1939--1948},
  year={2018},
  organization={PMLR},
  volume={80}
}

@misc{roth2022uncertain,
  author       = {Aaron Roth},
  title        = {Uncertain: Modern Topics in Uncertainty Quantification},
  year         = {2022},
  note         = {Textbook, University of Pennsylvania},
  howpublished = {\url{https://www.cis.upenn.edu/~aaroth/uncertain.html}},
}

@incollection{thiel2023generative,
  title={Generative ML and CSAM: Implications and mitigations},
  author={Thiel, David and Stroebel, Melissa and Portnoff, Rebecca},
  booktitle={Stanford digital repository},
  year={2023}
}

@misc{thornsafetybydesign,
title={Safety by Design for Generative AI: Preventing Child Sexual Abuse},
url={https://info.thorn.org/hubfs/thorn-safety-by-design-for-generative-AI.pdf},
author={Thorn and {All Tech is Human}},
year=2024
}

@article{kim2022universal,
  title={Universal Adaptability: Target-Independent Inference That Competes with Propensity Scoring},
  author={Kim, Michael P. and Kern, Christoph and Goldwasser, Shafi and Kreuter, Frauke and Reingold, Omer},
  journal={Proceedings of the National Academy of Sciences},
  volume={119},
  number={4},
  pages={e2108097119},
  year={2022},
  publisher={National Academy of Sciences},
  doi={10.1073/pnas.2108097119},
  url={https://www.pnas.org/doi/10.1073/pnas.2108097119}
}

@conference{kapoor2024societal,
  title={On the societal impact of open foundation models},
  author={Kapoor, Sayash and Bommasani, Rishi and Klyman, Kevin and Longpre, Shayne and Ramaswami, Ashwin and Cihon, Peter and Hopkins, Aspen and Bankston, Kevin and Biderman, Stella and Bogen, Miranda and others},
  booktitle={International Conference on Machine Learning, Position Paper Track},
  year={2024}
}

@inproceedings{yeom2018privacy,
  title={Privacy risk in machine learning: Analyzing the connection to overfitting},
  author={Yeom, Samuel and Giacomelli, Irene and Fredrikson, Matt and Jha, Somesh},
  booktitle={2018 IEEE 31st computer security foundations symposium (CSF)},
  pages={268--282},
  year={2018},
  organization={IEEE}
}

@article{thiel2023identifying,
  title={Identifying and eliminating csam in generative ml training data and models},
  author={Thiel, David},
  journal={Stanford Internet Observatory, Cyber Policy Center},
  volume={23},
  pages={3},
  year={2023}
}

@inproceedings{shokri2017membership,
  title={Membership inference attacks against machine learning models},
  author={Shokri, Reza and Stronati, Marco and Song, Congzheng and Shmatikov, Vitaly},
  booktitle={2017 IEEE symposium on security and privacy (SP)},
  pages={3--18},
  year={2017},
  organization={IEEE}
}

@inproceedings{song2019privacy,
  title={Privacy risks of securing machine learning models against adversarial examples},
  author={Song, Liwei and Shokri, Reza and Mittal, Prateek},
  booktitle={Proceedings of the 2019 ACM SIGSAC conference on computer and communications security},
  pages={241--257},
  year={2019}
}

@inproceedings{carlini2023extracting,
  title={Extracting training data from diffusion models},
  author={Carlini, Nicholas and Hayes, Jamie and Nasr, Milad and Jagielski, Matthew and Sehwag, Vikash and Tramer, Florian and Balle, Borja and Ippolito, Daphne and Wallace, Eric},
  booktitle={32nd USENIX Security Symposium (USENIX Security 23)},
  pages={5253--5270},
  year={2023}
}

@article{bertran2023scalable,
  title={Scalable membership inference attacks via quantile regression},
  author={Bertran, Martin and Tang, Shuai and Roth, Aaron and Kearns, Michael and Morgenstern, Jamie H and Wu, Steven Z},
  journal={Advances in Neural Information Processing Systems},
  volume={36},
  pages={314--330},
  year={2023}
}

@article{tang2023membership,
  title={Membership inference attacks on diffusion models via quantile regression},
  author={Tang, Shuai and Wu, Zhiwei Steven and Aydore, Sergul and Kearns, Michael and Roth, Aaron},
  journal={arXiv preprint arXiv:2312.05140},
  year={2023}
}

@inproceedings{secmi,
  title={Are diffusion models vulnerable to membership inference attacks?},
  author={Duan, Jinhao and Kong, Fei and Wang, Shiqi and Shi, Xiaoshuang and Xu, Kaidi},
  booktitle={International Conference on Machine Learning},
  pages={8717--8730},
  year={2023},
  organization={PMLR}
}

@inproceedings{choquette2021label,
  title={Label-only membership inference attacks},
  author={Choquette-Choo, Christopher A and Tramer, Florian and Carlini, Nicholas and Papernot, Nicolas},
  booktitle={International conference on machine learning},
  pages={1964--1974},
  year={2021},
  organization={PMLR}
}

@article{zarifzadeh2023rmia,
  title={Low-cost high-power membership inference attacks},
  author={Zarifzadeh, Sajjad and Liu, Philippe and Shokri, Reza},
  journal={arXiv preprint arXiv:2312.03262},
  year={2023}
}

@inproceedings{yichuan2024assessing,
  title={Assessing Membership Inference Attacks under Distribution Shifts},
  author={Yichuan, Shi and Kotevska, Olivera and Reshniak, Viktor and Sadovnik, Amir},
  booktitle={2024 IEEE International Conference on Big Data (BigData)},
  pages={4127--4131},
  year={2024},
  organization={IEEE}
}

@inproceedings{liu2022membership,
  title={Membership inference attacks by exploiting loss trajectory},
  author={Liu, Yiyong and Zhao, Zhengyu and Backes, Michael and Zhang, Yang},
  booktitle={Proceedings of the 2022 ACM SIGSAC Conference on Computer and Communications Security},
  pages={2085--2098},
  year={2022}
}

@article{watson2021importance,
  title={On the importance of difficulty calibration in membership inference attacks},
  author={Watson, Lauren and Guo, Chuan and Cormode, Graham and Sablayrolles, Alex},
  journal={arXiv preprint arXiv:2111.08440},
  year={2021}
}

@article{zhang2024membership,
  title={Membership inference attacks cannot prove that a model was trained on your data},
  author={Zhang, Jie and Das, Debeshee and Kamath, Gautam and Tram{\`e}r, Florian},
  journal={arXiv preprint arXiv:2409.19798},
  year={2024}
}

@techreport{Radford2019Language,
  author      = {Alec Radford and Jeffrey Wu and Rewon Child and David Luan and Dario Amodei and Ilya Sutskever},
  title       = {Language Models are Unsupervised Multitask Learners},
  institution = {OpenAI},
  year        = {2019},
  month       = {February},
  note        = {OpenAI Blog}
}

@inproceedings{li2021membership,
  title={Membership leakage in label-only exposures},
  author={Li, Zheng and Zhang, Yang},
  booktitle={Proceedings of the 2021 ACM SIGSAC Conference on Computer and Communications Security},
  pages={880--895},
  year={2021}
}

@inproceedings{leino2020stolen,
  title={Stolen memories: Leveraging model memorization for calibrated $\{$White-Box$\}$ membership inference},
  author={Leino, Klas and Fredrikson, Matt},
  booktitle={29th USENIX security symposium (USENIX Security 20)},
  pages={1605--1622},
  year={2020}
}

@article{steed2024quantifying,
  title={Quantifying privacy risks of public statistics to residents of subsidized housing},
  author={Steed, Ryan and Qing, Diana and Wu, Zhiwei Steven},
  journal={arXiv preprint arXiv:2407.04776},
  year={2024}
}

@article{tramer2022position,
  title={Position: Considerations for differentially private learning with large-scale public pretraining},
  author={Tram{\`e}r, Florian and Kamath, Gautam and Carlini, Nicholas},
  journal={arXiv preprint arXiv:2212.06470},
  year={2022}
}

@article{loshchilov2017decoupled,
  title={Decoupled weight decay regularization},
  author={Loshchilov, Ilya and Hutter, Frank},
  journal={arXiv preprint arXiv:1711.05101},
  year={2017}
}

@article{hayes2025strong,
  title={Strong Membership Inference Attacks on Massive Datasets and (Moderately) Large Language Models},
  author={Hayes, Jamie and Shumailov, Ilia and Choquette-Choo, Christopher A and Jagielski, Matthew and Kaissis, George and Lee, Katherine and Nasr, Milad and Ghalebikesabi, Sahra and Mireshghallah, Niloofar and Annamalai, Meenatchi Sundaram Mutu Selva and others},
  journal={arXiv preprint arXiv:2505.18773},
  year={2025}
}

@inproceedings{nasr2019comprehensive,
  title={Comprehensive privacy analysis of deep learning: Passive and active white-box inference attacks against centralized and federated learning},
  author={Nasr, Milad and Shokri, Reza and Houmansadr, Amir},
  booktitle={2019 IEEE Symposium on Security and Privacy (SP)},
  pages={739--753},
  year={2019},
  organization={IEEE}
}

@inproceedings{ye2022enhanced,
  title={Enhanced membership inference attacks against machine learning models},
  author={Ye, J. and Maddi, A. and Murakonda, S. K. and Bindschaedler, V. and Shokri, Reza},
  booktitle={Proceedings of the 2022 ACM SIGSAC Conference on Computer and Communications Security},
  pages={3093--3106},
  year={2022}
}

@article{mattern2023membership,
  title={Membership inference attacks against language models via neighbourhood comparison},
  author={Mattern, J. and Mireshghallah, F. and Jin, Z. and Sch{\"o}lkopf, B. and Sachan, M. and Berg-Kirkpatrick, T.},
  journal={arXiv preprint arXiv:2305.18462},
  year={2023}
}

@article{shi2023detecting,
  title={Detecting pretraining data from large language models},
  author={Shi, Weijia and Ajith, A. and Xia, Mengzhou and Huang, Yukun and Liu, D. and Blevins, Terra and Zettlemoyer, Luke},
  journal={arXiv preprint arXiv:2310.16789},
  year={2023}
}
\bibliographystyle{icml2026}


\newpage
\appendix
\onecolumn
\appendix
\section{Experiments and Implementation Details}
\label{appx:implementation-details}

Shadow models were trained on 2 NVIDIA A100 GPUs using the code released by ~\citet{lira}. 
The code was used unmodified except to drop the relevant classes from the attack model training data. 

The quantile regression models were trained on 4 A800 nodes, taking about 3000MB per node to train ConvNext-Tiny-224 models. Each model took as input a 224x224x3 image and returned 2 outputs, the predicted Gaussian mean and variance. Each model was trained for 30 epochs. For CINIC-10, training with no class dropout took approximately 40 minutes per model. For CIFAR-100, training took approximately 12 minutes per model. For ImageNet, training took approximately 5 hours per model. Under class dropout and sample dropout, the training dataset was smaller and training time reduced.

We also experimented with running ConvNext-Large-224, which took 11000MB per node to train, and significantly longer, i.e. 4 hours per model for CINIC-10. 

For completeness, below we also include the member and nonmember query set sizes for each dataset.
\begin{table}[t]
\centering
\begin{tabular}{lcc}
\toprule
\textbf{Dataset / reported setting} & \textbf{Average member query set size} & \textbf{Non-member query set size} \\
\midrule
\texttt{CINIC-10}, per dropped class & 8,979 & 9,000 \\
\texttt{CIFAR-100} (20 classes), per dropped superclass & 1,245 & 500 \\
\texttt{ImageNet}, full-distribution ROC & 480,868 & 320,175 \\
\texttt{ImageNet}, dropped-class-only ROC & $\sim 481 \times n_{\text{classes\_dropped}}$ & $\sim 320 \times n_{\text{classes\_dropped}}$ \\
\texttt{Texas}, per dropped 10-class bin & 2,701 & 1,351 \\
\texttt{20 Newsgroups}, per reported dropped group & 1,426 & 1,883 \\
\bottomrule
\end{tabular}
\caption{Query set sizes for member and non-member datasets across reported experimental settings.}
\label{tab:query_set_sizes}
\end{table}

\section{Additional Shadow Model Results}
\label{appx:shadowmodel-results}
We provide additional results in the FPR 0.1\% regime to supplement Figure \ref{fig:shadow-degradation}.

\begin{figure}[h!]
\centering
\begin{subfigure}{0.48\textwidth}
\includegraphics[width=\textwidth]{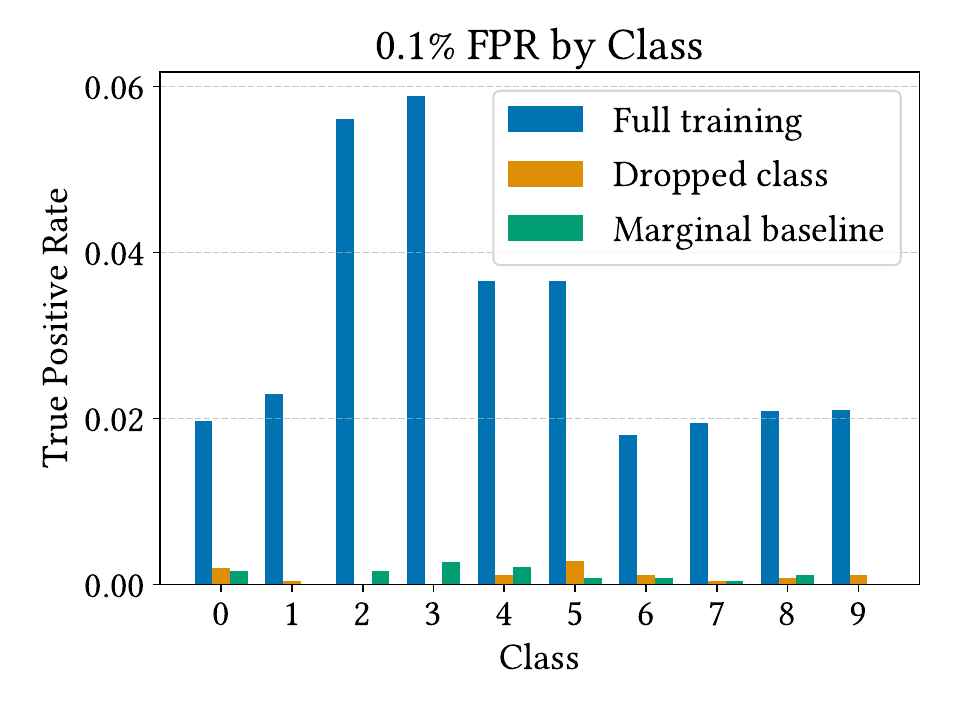}
\caption{Results for CINIC-10.}
\end{subfigure}
\begin{subfigure}{0.48\textwidth}
\includegraphics[width=\textwidth]{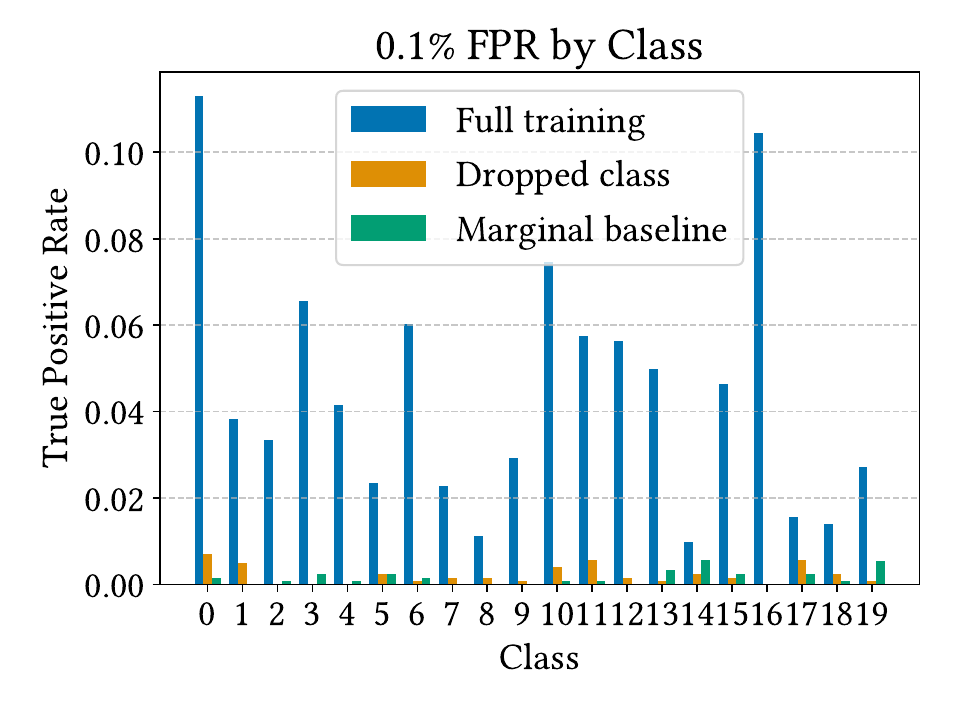}
\caption{Results for CIFAR-100 (coarse labels).
}
\end{subfigure}
\caption{True positive rates for shadow model attacks in the 0.1\% false positive rate regime for CINIC-10 and CIFAR-100. Each bar represents the TPR on the indicated class. In yellow, we plot the TPR when that class is excluded from shadow model training. The attack success degrades significantly under class exclusion, often performing worse than the marginal baseline (global threshold).}
\label{fig:shadow-degradation-2}
\end{figure}

As described in Section \ref{sec:expts-shadow}, one possible explanation for shadow models' underperformance is due to the true logit score function which fails in an unseen class setting. However, we also test the top-two logit difference as a score function, and find that shadow models perform even worse in this setting.

\begin{figure}[h!]
\centering
\includegraphics[width=0.7\textwidth]{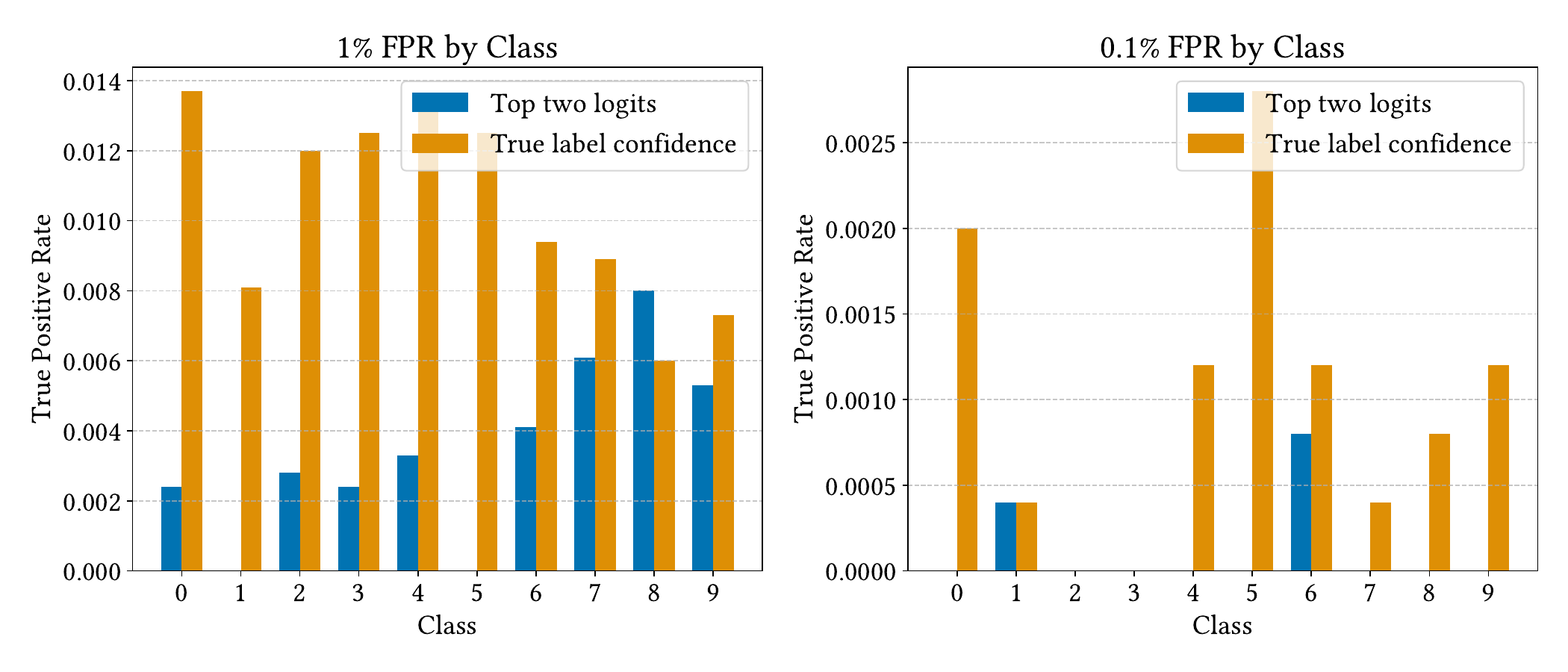}
        \setlength{\belowcaptionskip}{-1em}

\caption{Comparison between shadow model attack success (with class $i$ dropped) with true label confidence as the score metric and top-two logit difference as the score metric on CINIC-10. Although the top-two logit difference does not use the true (dropped) label that was not seen by the shadow models, the attack performs even worse than with the true label confidence.}
\label{fig:toptwo}
\end{figure}

\clearpage
\section{AUC Results}
\label{appx:auc-results}

We provide AUC results to supplement the TPR at low FPR results provided in the paper.
However, we note that AUC is \emph{not} the best-practice metric
for measuring MIA success (see, e.g., ~\citet{lira})
because AUC integrates over all false-positive rates.
Quoting ~\citet{lira}: ``[$\ldots$] the AUC is not
an appropriate measure of an attack’s efficacy, since the AUC
averages over all false-positive rates, including high error rates
that are irrelevant for a practical attack. The TPR of an attack
when the FPR is above 50\% is not meaningfully useful, yet
this regime accounts for more than half of its AUC score.''
Nevertheless, we provide AUC results for completeness here.

We find that results are inconclusive for AUC as compared to the low-FPR regime, where quantile regression definitively wins across datasets and domains. 
On image datasets, RMIA outperforms qunatile regression on CIFAR-10,
but results are inconclusive on CIFAR-100.
On the Texas tabular dataset and 
20 Newsgroups text dataset, quantile regression outperforms all other methods.

\begin{figure}[h!]
    \centering
    \begin{subfigure}[b]{0.5\textwidth}
        \centering
        \includegraphics[width=\textwidth]{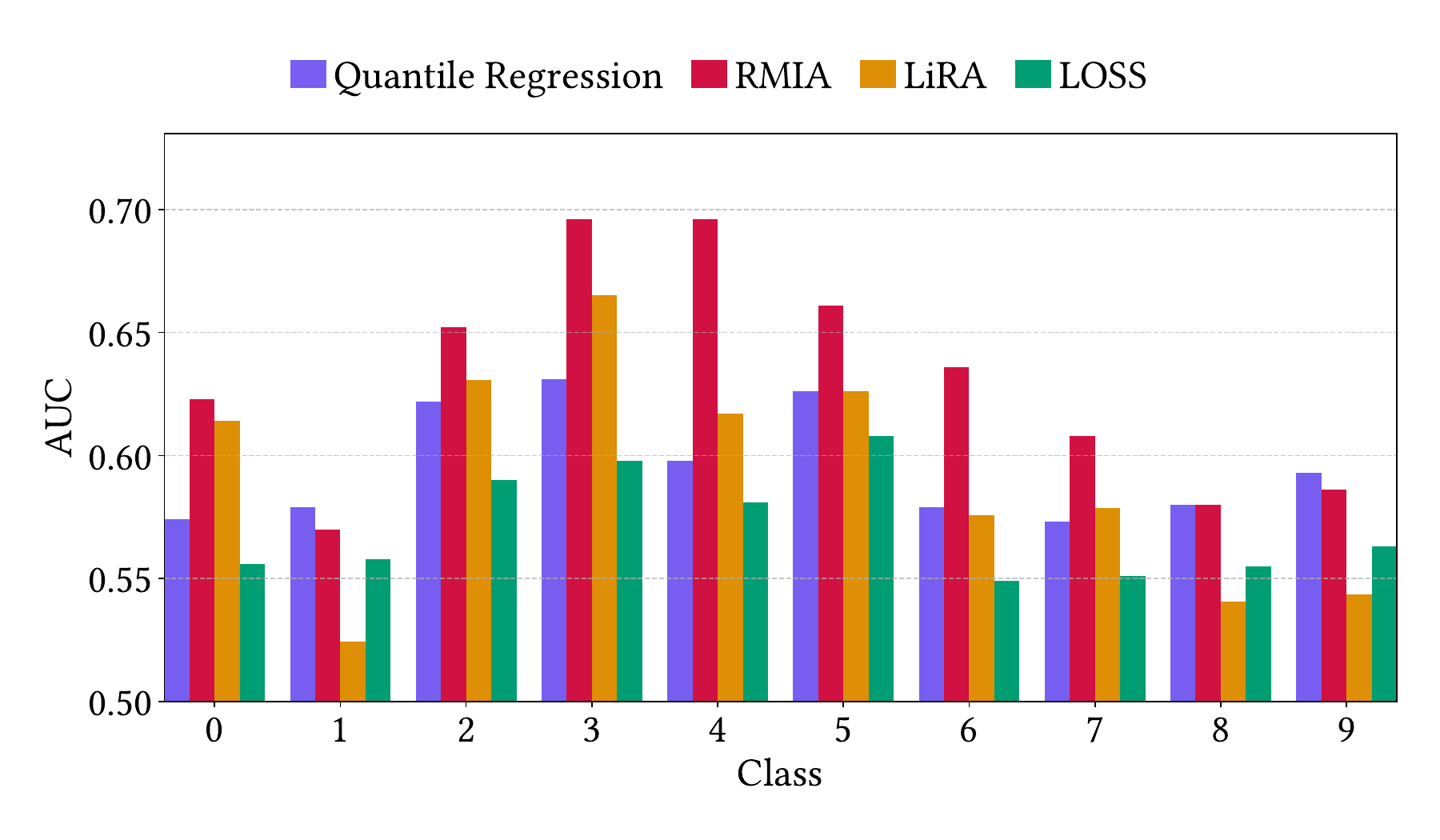} 
        \vspace{-.25in}
        \caption{AUC for the dropped out class, CIFAR-10}
        \label{fig:qr_cifar10_auc}
    \end{subfigure}\hfill%
    \begin{subfigure}[b]{0.5\textwidth}
        \centering
        \includegraphics[width=\textwidth]{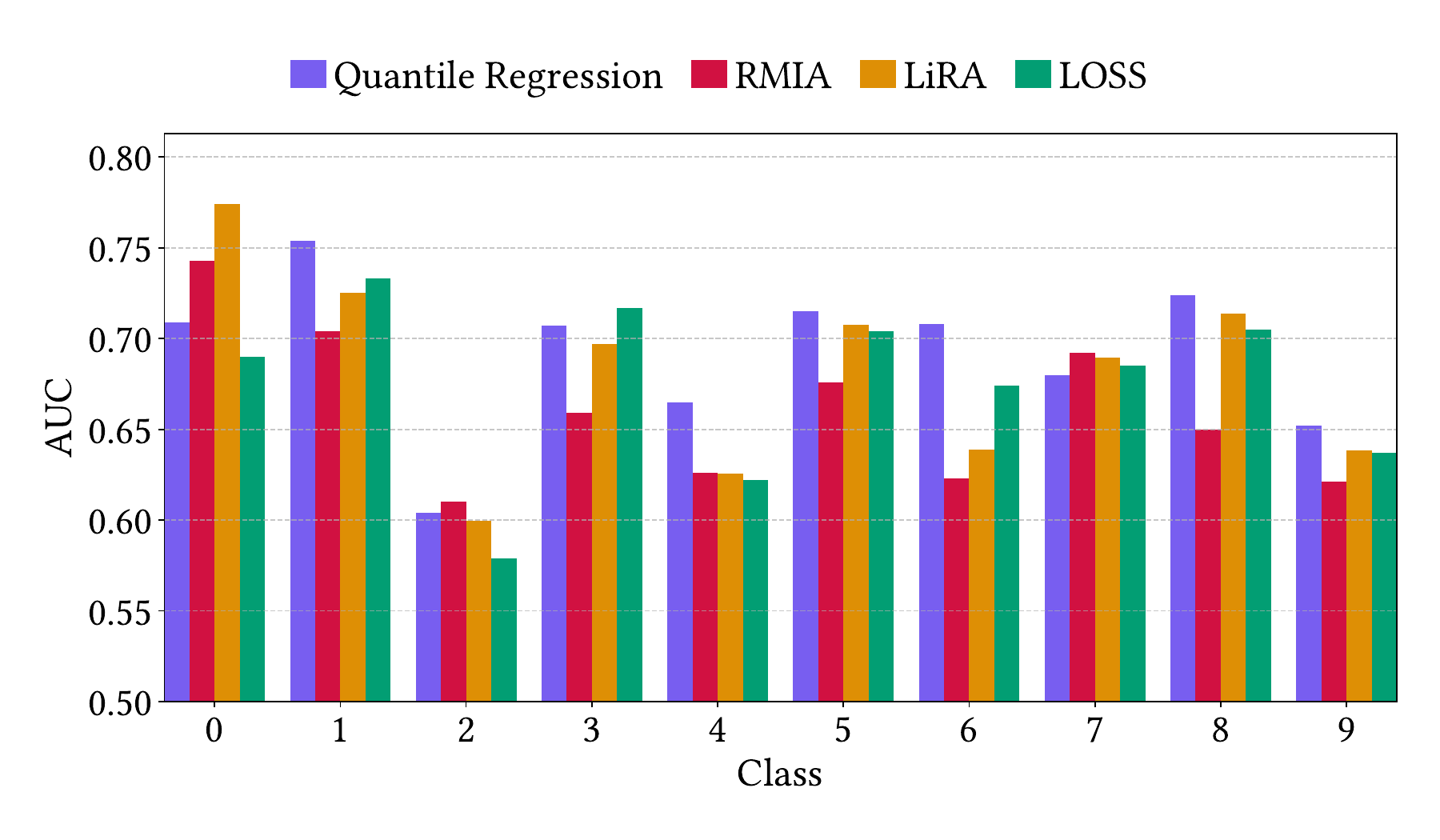} 
        \vspace{-.25in}
        \caption{AUC for the dropped out superclass, CIFAR-100.}
        \label{fig:qr_cifar100_auc}
    \end{subfigure}
    \newline
    \begin{subfigure}[b]{0.5\textwidth}
        \centering
        \includegraphics[width=\textwidth]{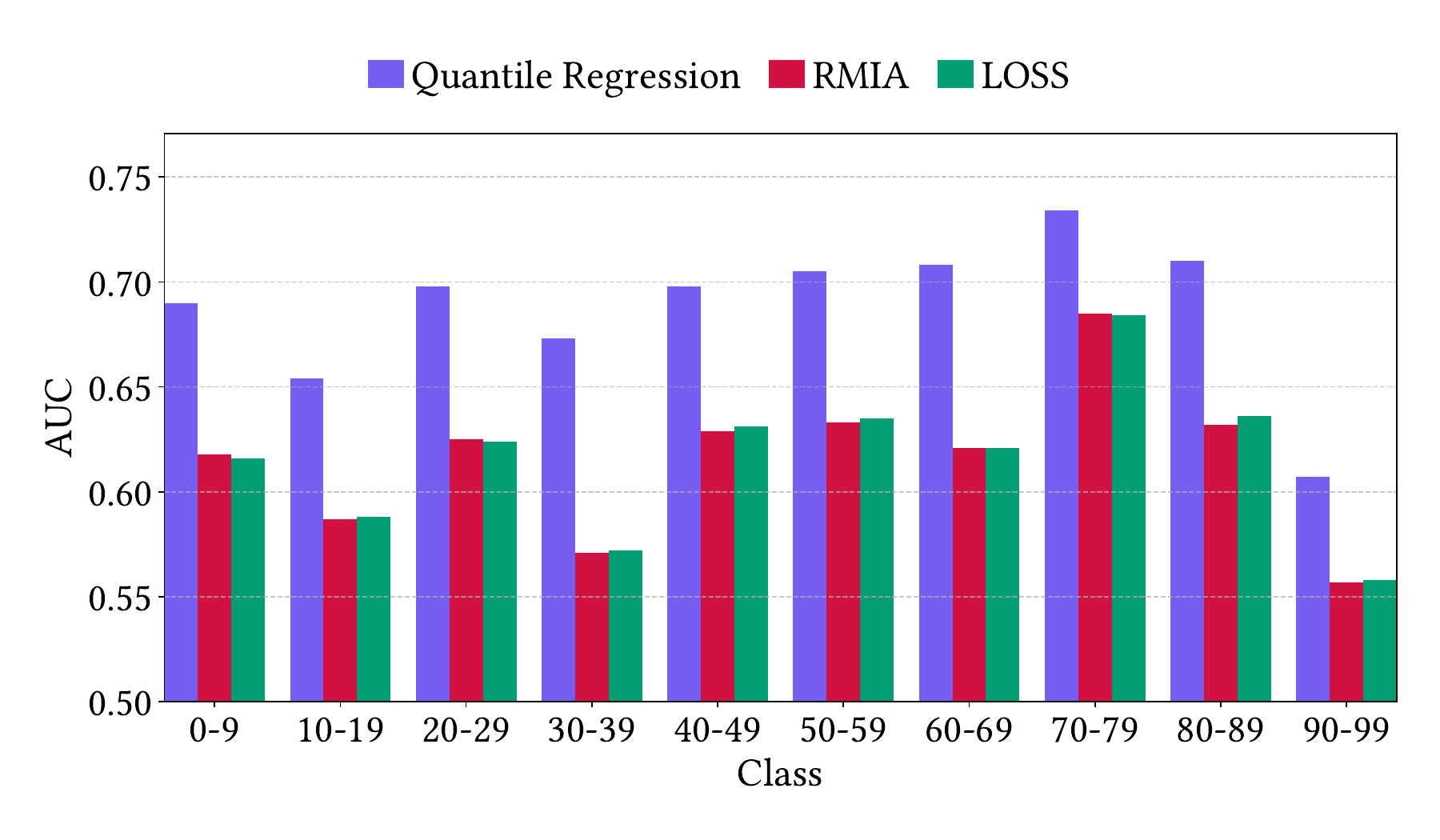} 
        \vspace{-.25in}
        \caption{AUC for the dropped out class, Texas}
        \label{fig:qr_texas_auc}
    \end{subfigure}\hfill%
    \begin{subfigure}[b]{0.5\textwidth}
        \centering
        \includegraphics[width=\textwidth]{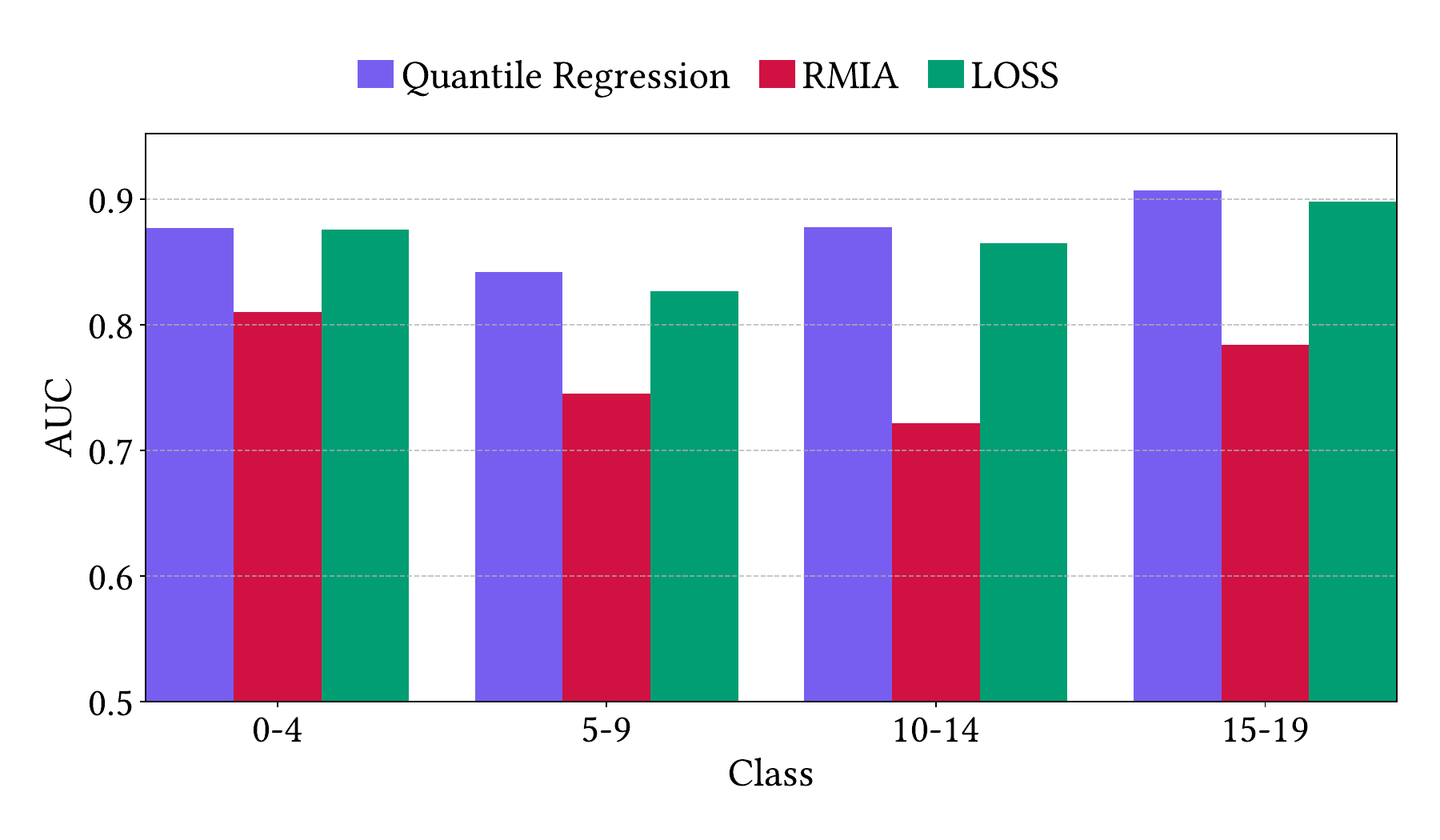} 
        \vspace{-.25in}
        \caption{AUC for the dropped out superclass, 20 Newsgroups.}
        \label{fig:qr_ng20_auc}
    \end{subfigure}
    \caption{AUCs for CINIC-10, CIFAR-100, Texas (tabular) and 20 Newsgroups (text) on sets of unseen classes. Each bar represents the AUC on classes $C$ when $C$ are dropped from the attack training set.}
    \label{fig:appx:qr_aucs}
    \vspace{-.15in}
\end{figure}

In addition, to mitigate the effects of integrating over all 
FPRs, we provide \emph{truncated} AUC results,
which provide a middle ground between pointwise low-FPR results
and full AUC.

\begin{figure}[h!]
    \centering
    \begin{subfigure}[b]{0.5\textwidth}
        \centering
        \includegraphics[width=\textwidth]{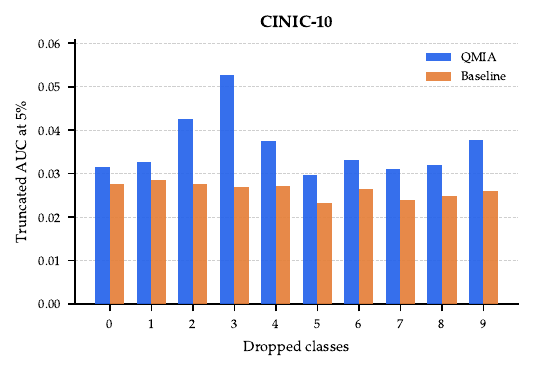} 
        \vspace{-.25in}
        \caption{Truncated AUC for the dropped out class, CIFAR-10}
        \label{fig:qr_cifar10_auc}
    \end{subfigure}\hfill%
    \begin{subfigure}[b]{0.5\textwidth}
        \centering
        \includegraphics[width=\textwidth]{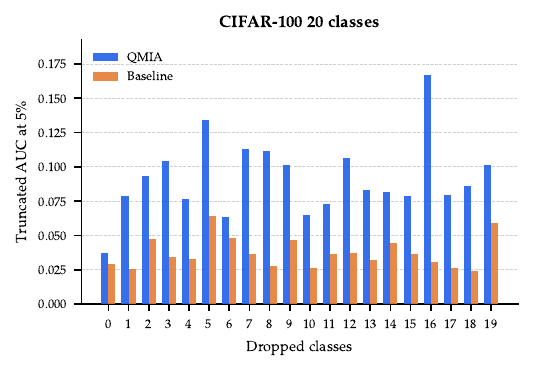} 
        \vspace{-.25in}
        \caption{Truncated AUC for the dropped out superclass, CIFAR-100.}
        \label{fig:qr_cifar100_auc}
    \end{subfigure}
    \newline
    \begin{subfigure}[b]{0.5\textwidth}
        \centering
        \includegraphics[width=\textwidth]{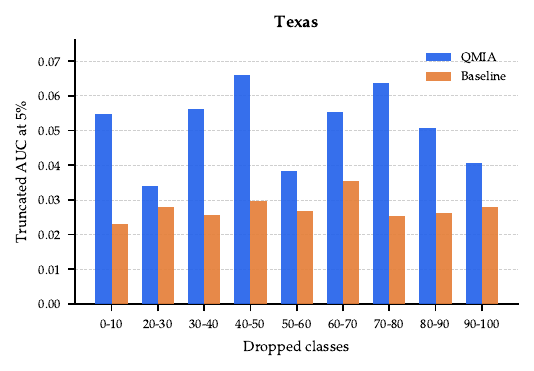} 
        \vspace{-.25in}
        \caption{Truncated AUC for the dropped out class, Texas.}
        \label{fig:qr_texas_auc}
    \end{subfigure}\hfill%
    \begin{subfigure}[b]{0.5\textwidth}
        \centering
        \includegraphics[width=\textwidth]{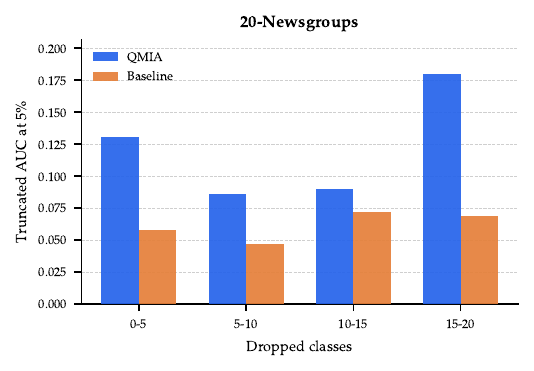} 
        \vspace{-.25in}
        \caption{Truncated AUC for the dropped out superclass, 20 Newsgroups.}
        \label{fig:qr_ng20_auc}
    \end{subfigure}
    \caption{AUCs truncated at 5\% FPR for CINIC-10, CIFAR-100, Texas (tabular) and 20 Newsgroups (text) on sets of unseen classes. Each bar represents the truncated AUC on classes $C$ when $C$ are dropped from the attack training set.}
    \label{fig:appx:qr_aucs}
    \vspace{-.15in}
\end{figure}

\clearpage
\section{CINIC-10 and CIFAR-100 Multiclass Dropout Results}
\label{appx:multiclass-drop-results}

We additionally provide AUC and TPR at FPR=1\% for settings in CINIC-10 and CIFAR-100 where multiple classes are dropped. In this setting, quantile regression continues to outperform the LOSS baseline, achieving TPR of ~3\% at low FPR even when half of the superclasses are dropped from CIFAR-100.
AUC scores are comparable or better for quantile regression and LOSS.

\begin{figure}[h!]
    \centering
    \begin{subfigure}[b]{0.5\textwidth}
        \centering
        \includegraphics[width=\textwidth]{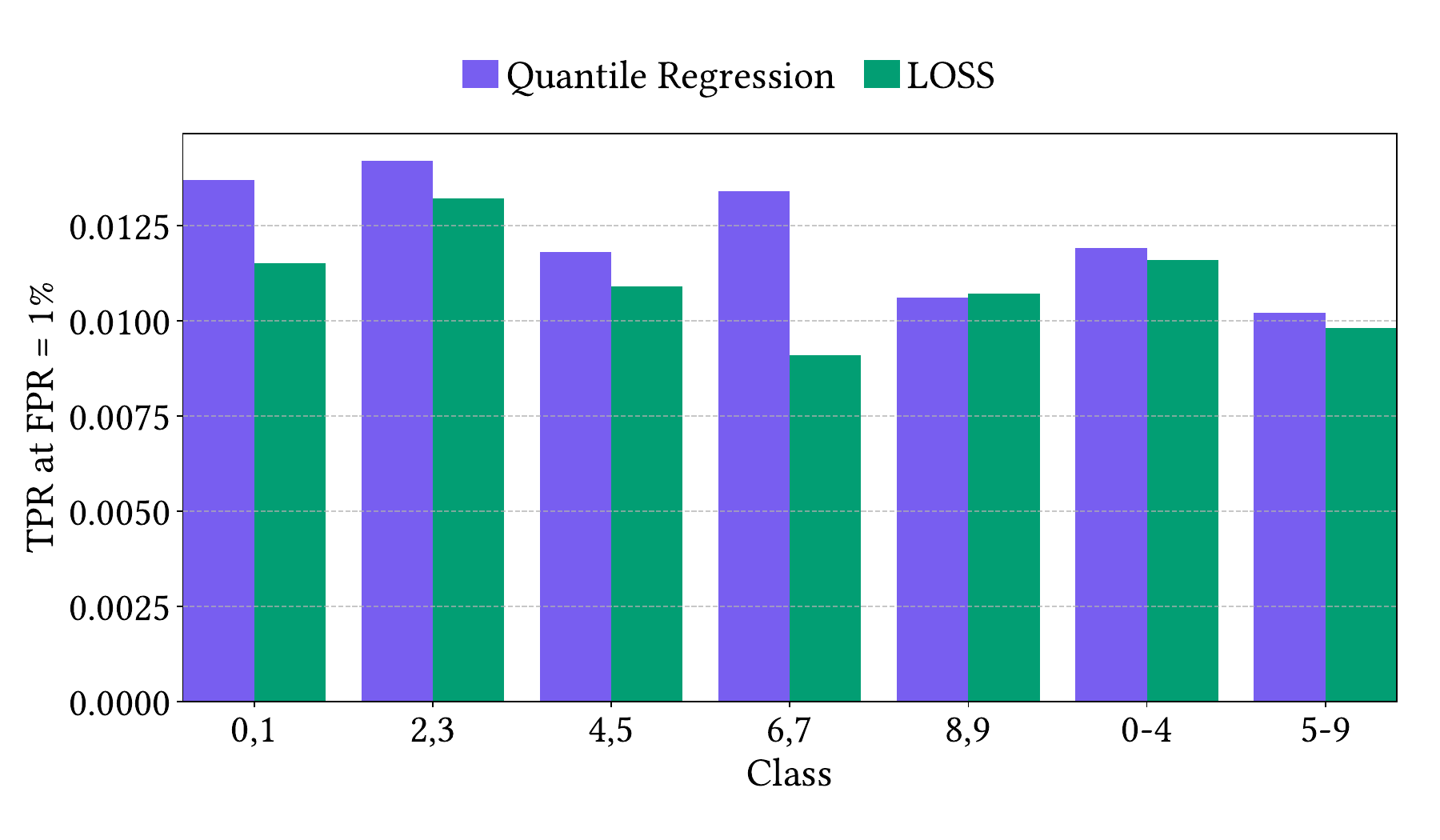} 
        \vspace{-.25in}
        \caption{TPR for the dropped out classes, CIFAR-10.}
        \label{fig:qr_cifar10_fpr1_multidrop}
    \end{subfigure}\hfill%
    \begin{subfigure}[b]{0.5\textwidth}
        \centering
        \includegraphics[width=\textwidth]{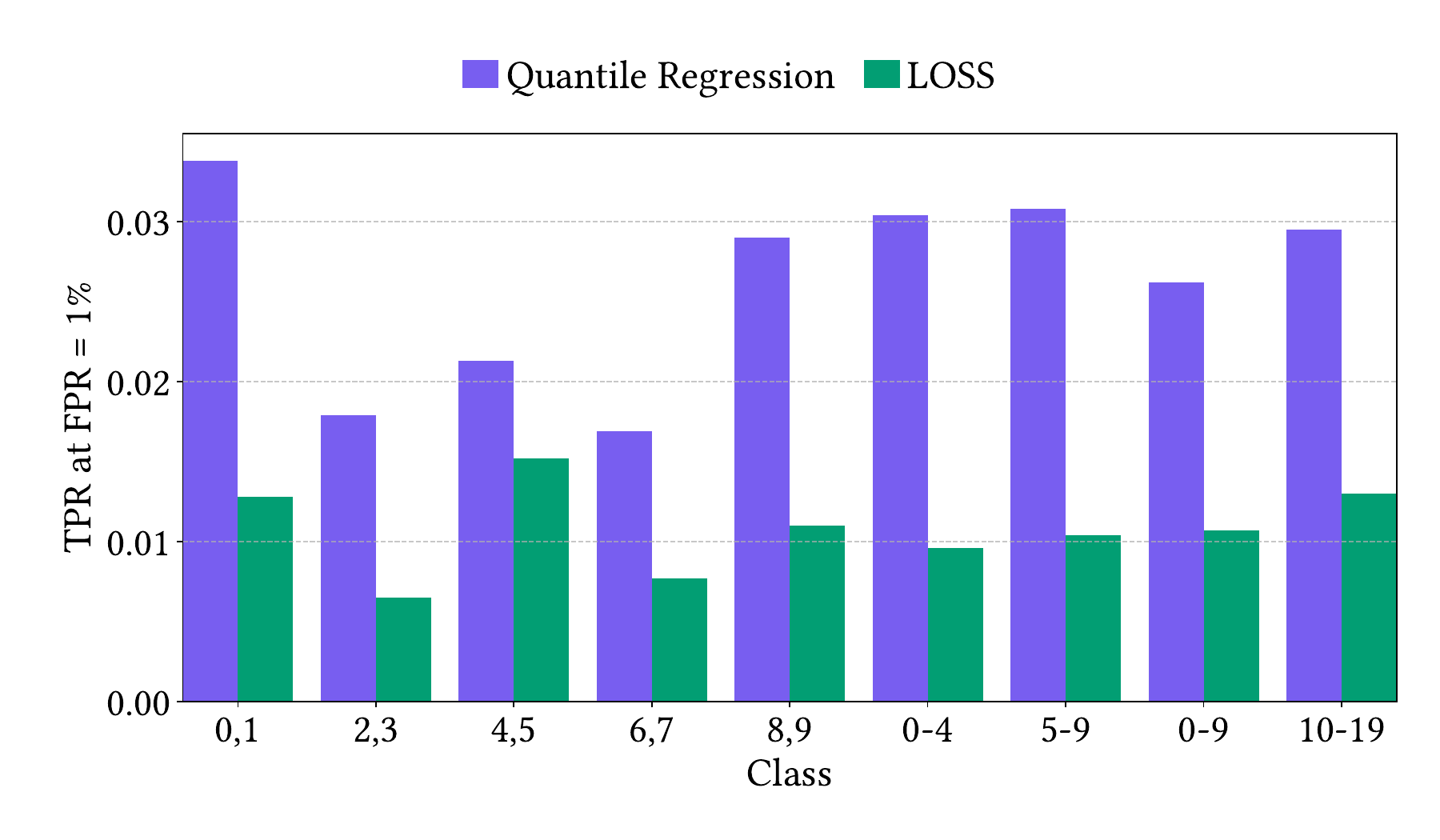} 
        \vspace{-.25in}
        \caption{TPR for the dropped out superclasses, CIFAR-100.}
        \label{fig:qr_cifar100_fpr1_multidrop}
    \end{subfigure}
    \caption{TPRs for CINIC-10 and CIFAR-100 on sets of unseen classes. Each bar represents the TPR on classes $C$ when $C$ are dropped from the attack training set. We only report results at 1\% FPR; the results at 0.1\% FPR are not meaningful due to the small sample size of the validation set on a single class.}
    \label{fig:appx:multidrop_tprs}
    \vspace{-.15in}
\end{figure}

\begin{figure}[h!]
    \centering
    \begin{subfigure}[b]{0.5\textwidth}
        \centering
        \includegraphics[width=\textwidth]{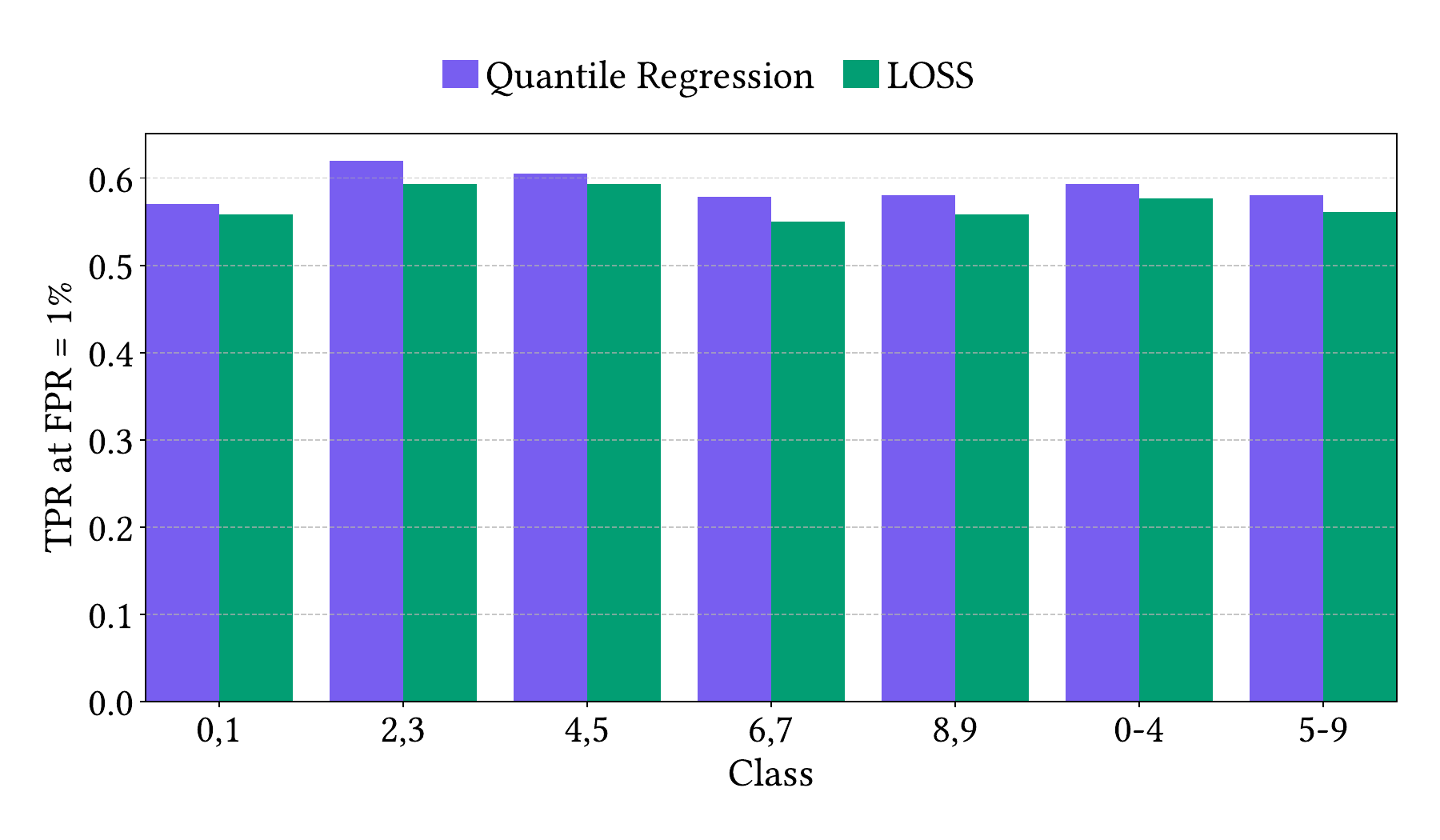} 
        \vspace{-.25in}
        \caption{AUC for the dropped out classes, CINIC-10.}
        \label{fig:qr_cifar10_auc_multidrop}
    \end{subfigure}\hfill%
    \begin{subfigure}[b]{0.5\textwidth}
        \centering
        \includegraphics[width=\textwidth]{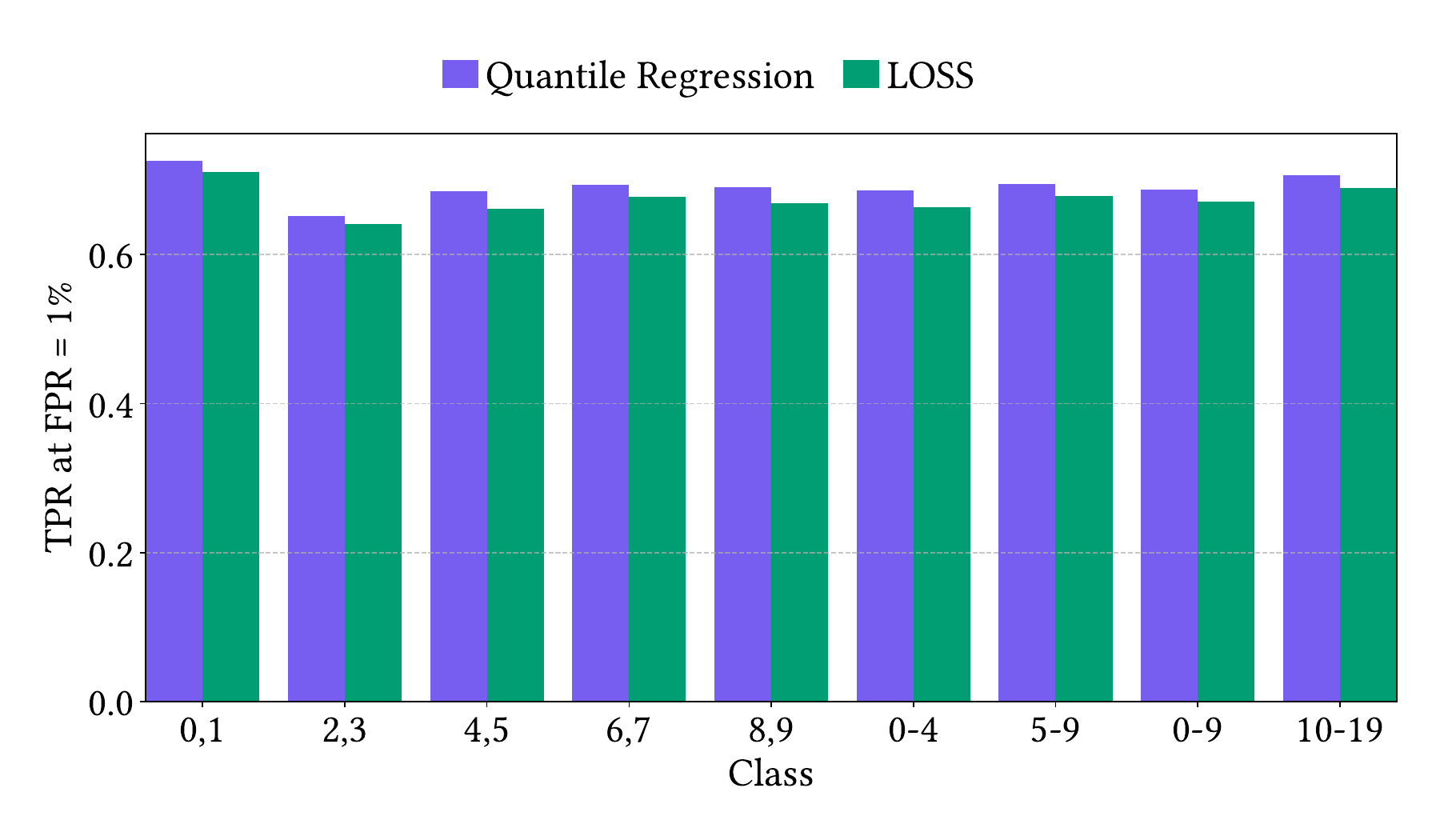} 
        \vspace{-.25in}
        \caption{AUC for the dropped out superclasses, CIFAR-100.}
        \label{fig:qr_cifar100_auc_multidrop}
    \end{subfigure}
    \caption{AUCs for CINIC-10 and CIFAR-100 on sets of unseen classes. Each bar represents the AUC on classes $C$ when $C$ are dropped from the attack training set.}
    \label{fig:appx:multidrop_aucs}
    \vspace{-.15in}
\end{figure}

\section{ROC Curves}
\label{appx:roc}

In addition to the pointwise low-FPR results in the main text
and the AUC results above, for completeness we also include the full ROC results
for our main class-drop experiments below. 

\begin{figure}[t]
    \centering
    \begin{subfigure}[b]{0.49\linewidth}
        \centering
        \includegraphics[width=\linewidth]{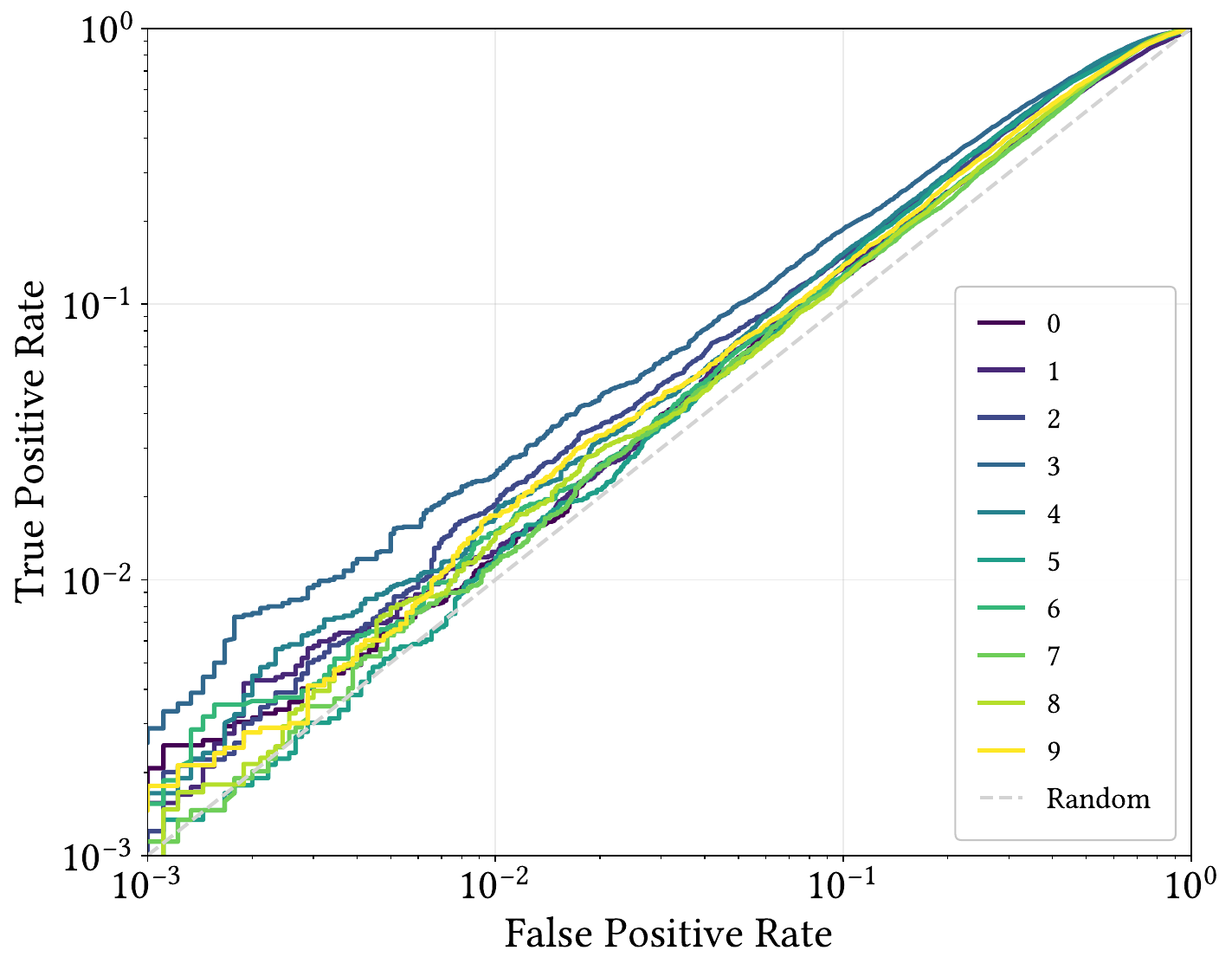}
        \caption{CINIC-10.}
        \label{fig:roc_cinic10}
    \end{subfigure}
    \hfill
    \begin{subfigure}[b]{0.49\linewidth}
        \centering
        \includegraphics[width=\linewidth]{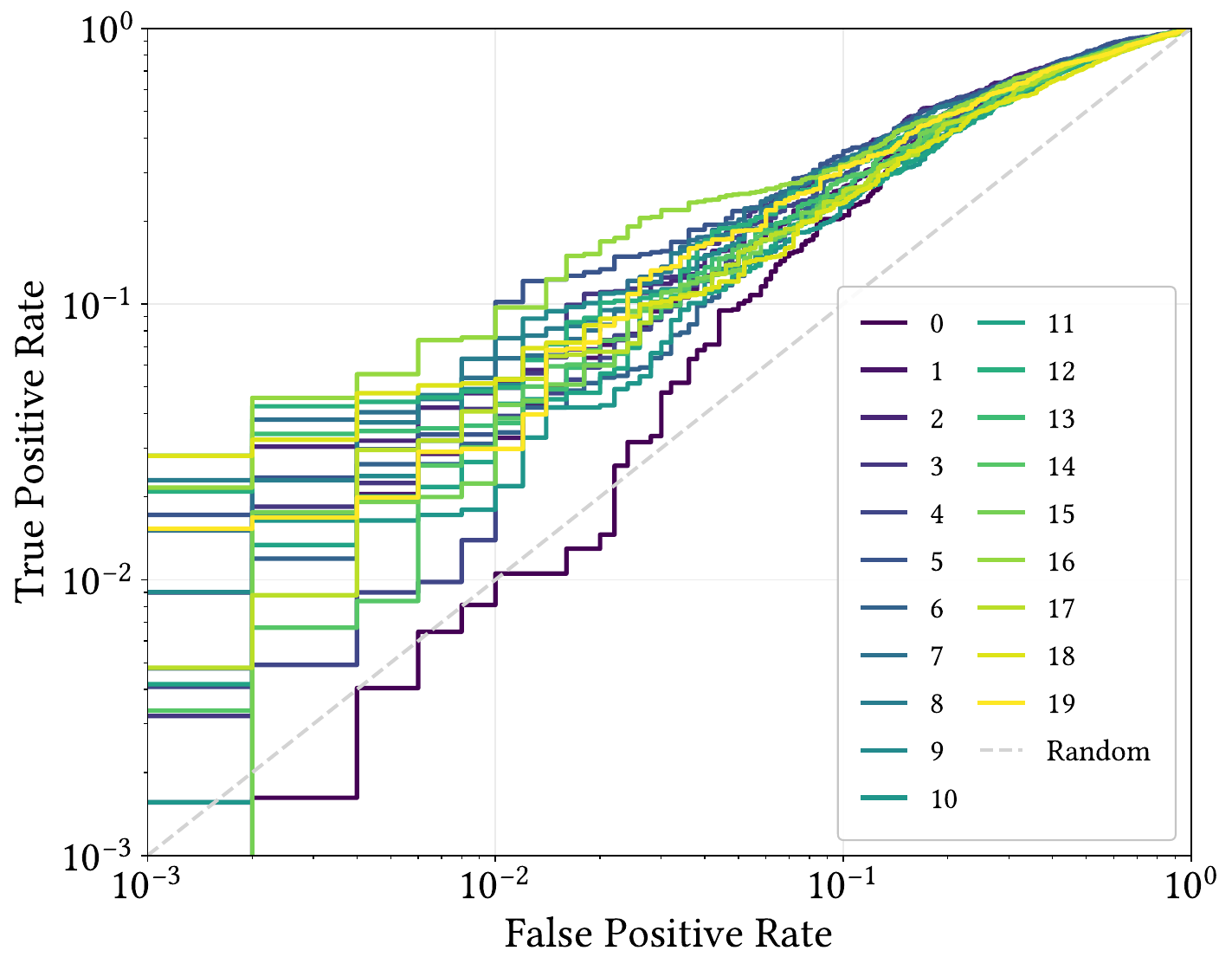}
        \caption{CIFAR-100 (superclass set).}
        \label{fig:roc_cifar100}
    \end{subfigure}
 
    \vspace{0.5em}
 
    \begin{subfigure}[b]{0.49\linewidth}
        \centering
        \includegraphics[width=\linewidth]{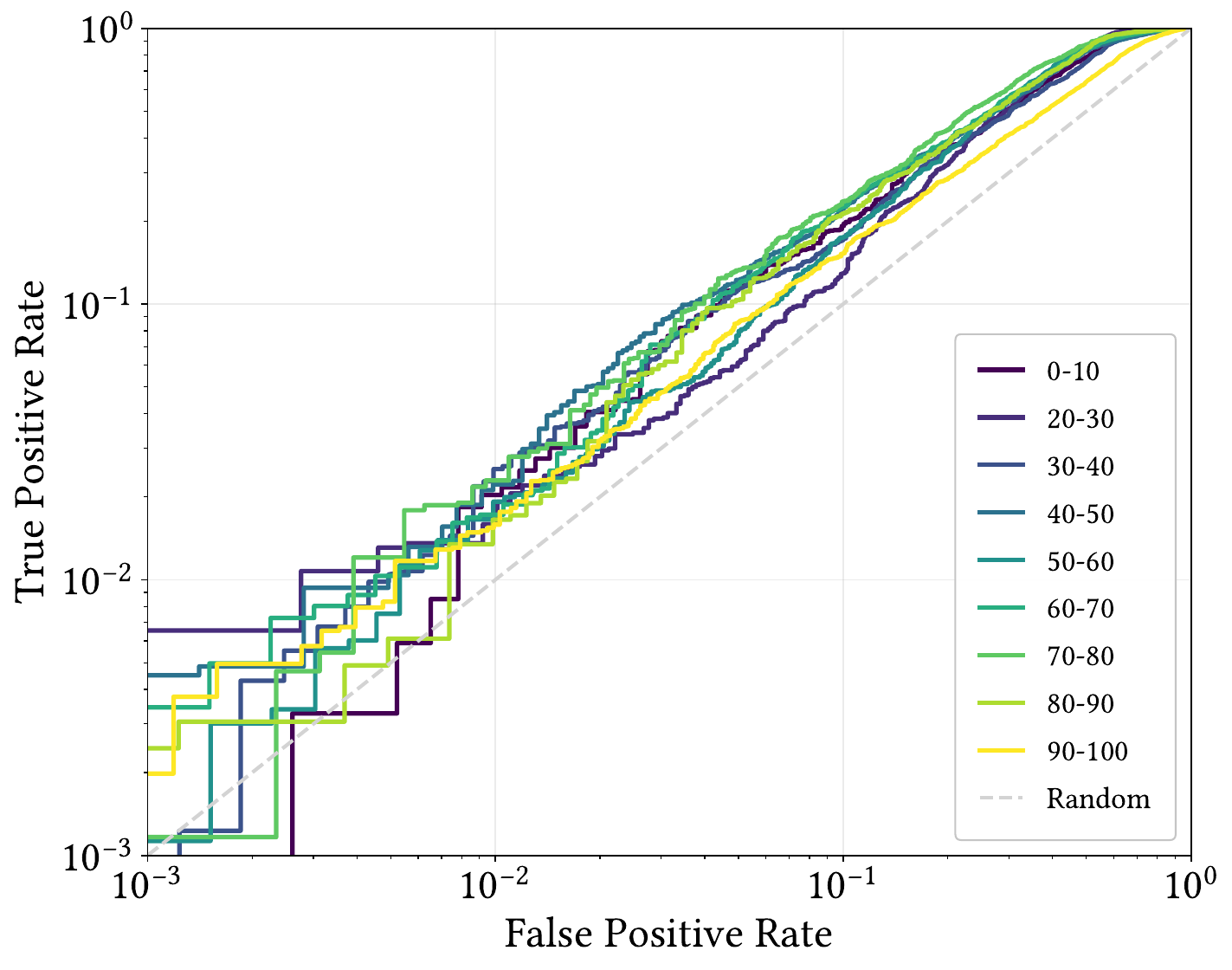}
        \caption{Texas.}
        \label{fig:roc_texas}
    \end{subfigure}
    \hfill
    \begin{subfigure}[b]{0.49\linewidth}
        \centering
        \includegraphics[width=\linewidth]{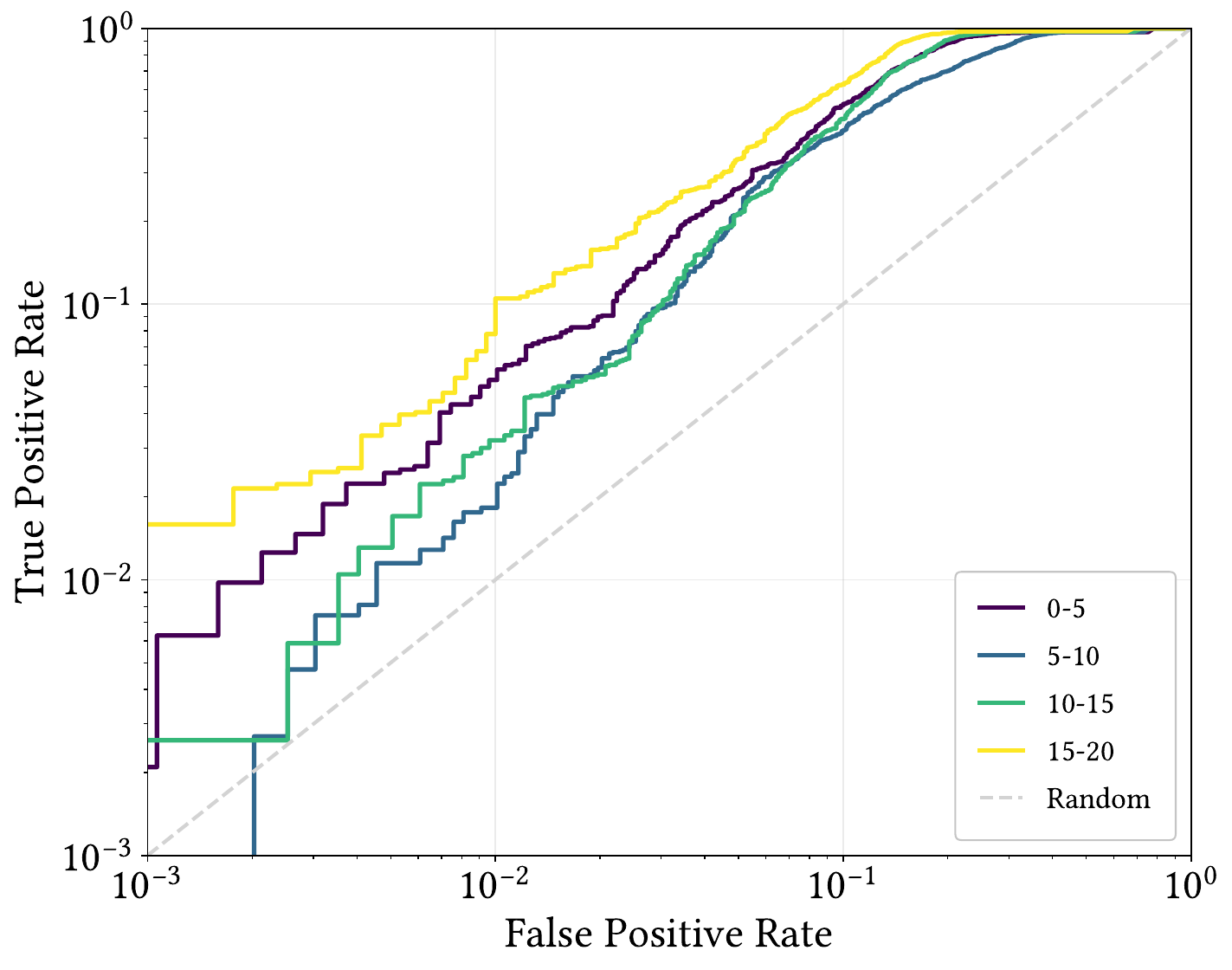}
        \caption{20-Newsgroups.}
        \label{fig:roc_newsgroups}
    \end{subfigure}
 
    \caption{ROC curves for CINIC-10, CIFAR-100 (superclass set), Texas, and 20-Newsgroups. Each line corresponds to the 
    ROC on the dropped class set. }
    \label{fig:roc_overlay}
\end{figure}

\clearpage
\section{CINIC-10 and CIFAR-100 Alternate Backbone Results}
\label{appx:image-architecture-results}

The CINIC-10 and CIFAR-100 in the main paper audit a ResNet-50 architecture classifier, so here we show quantile regression results for ResNet-18 and Vision Transformer (\texttt{vit-base-patch16-224}).
Again, quantile regression has a striking advantage at low FPR
while AUC scores are comparable or better across the board.

\begin{figure}[h!]
    \centering
    \begin{subfigure}[b]{0.5\textwidth}
        \centering
        \includegraphics[width=\textwidth]{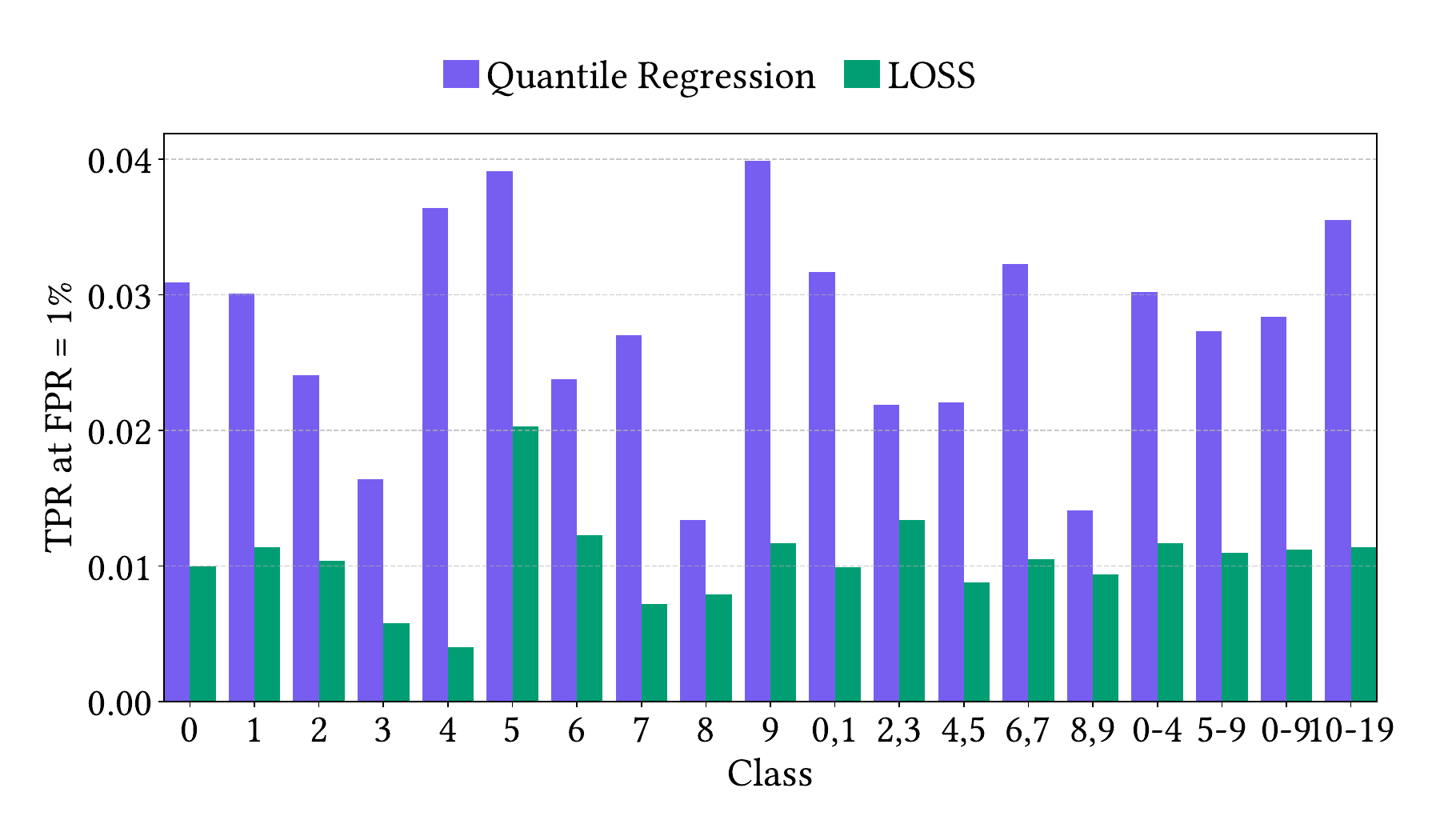} 
        \vspace{-.25in}
        \caption{TPR, CIFAR-100, ViT.}
        \label{fig:qr_cifar100_fpr1_vit}
    \end{subfigure}\hfill%
    \begin{subfigure}[b]{0.5\textwidth}
        \centering
        \includegraphics[width=\textwidth]{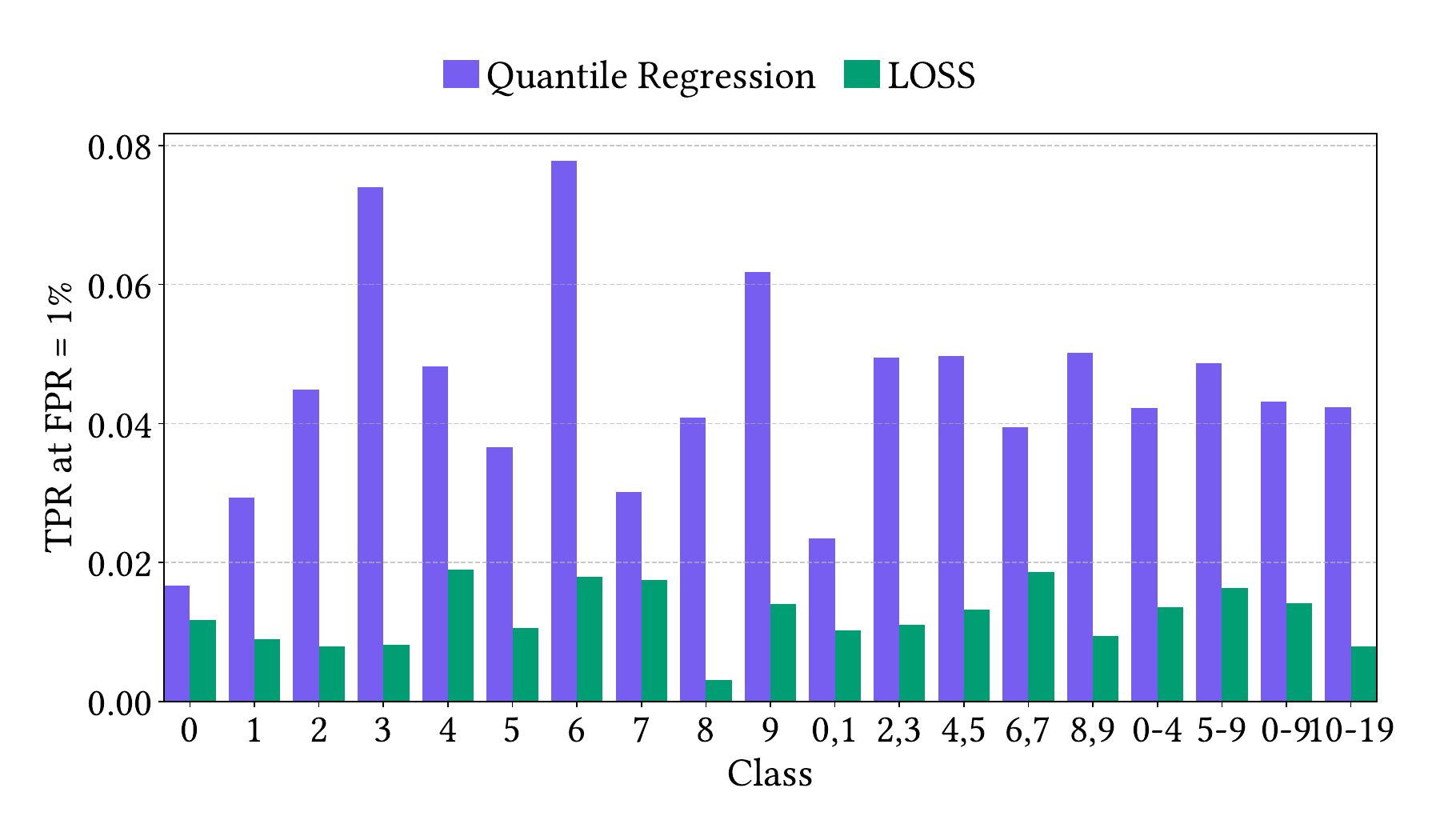} 
        \vspace{-.25in}
        \caption{TPR, CIFAR-100, ResNet-18.}
        \label{fig:qr_cifar100_fpr1_resnet18}
    \end{subfigure}
    \newline
    \begin{subfigure}[b]{0.5\textwidth}
        \centering
        \includegraphics[width=\textwidth]{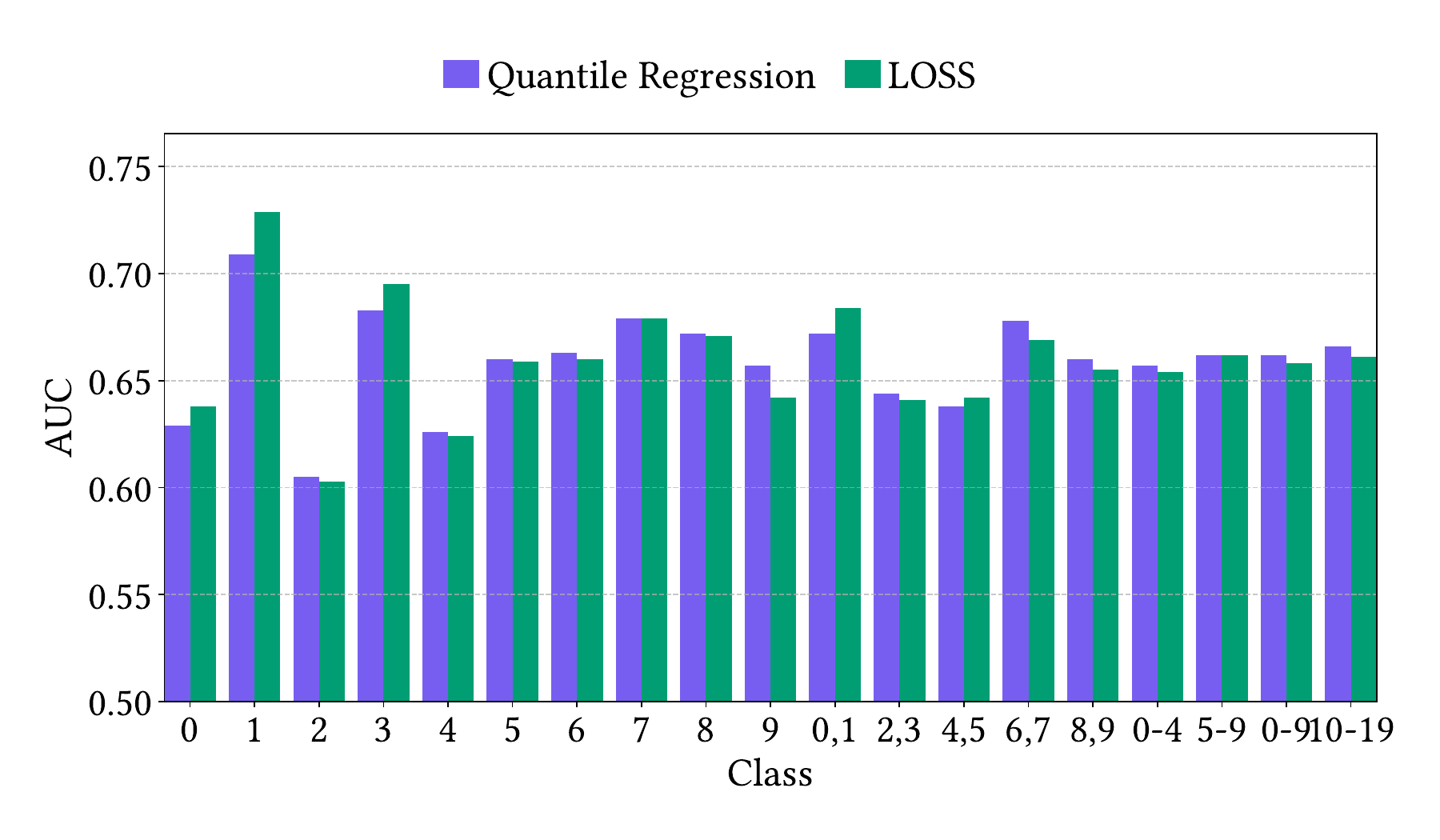} 
        \vspace{-.25in}
        \caption{AUC, CIFAR-100, ViT.}
        \label{fig:qr_cifar100_auc_vit}
    \end{subfigure}\hfill%
    \begin{subfigure}[b]{0.5\textwidth}
        \centering
        \includegraphics[width=\textwidth]{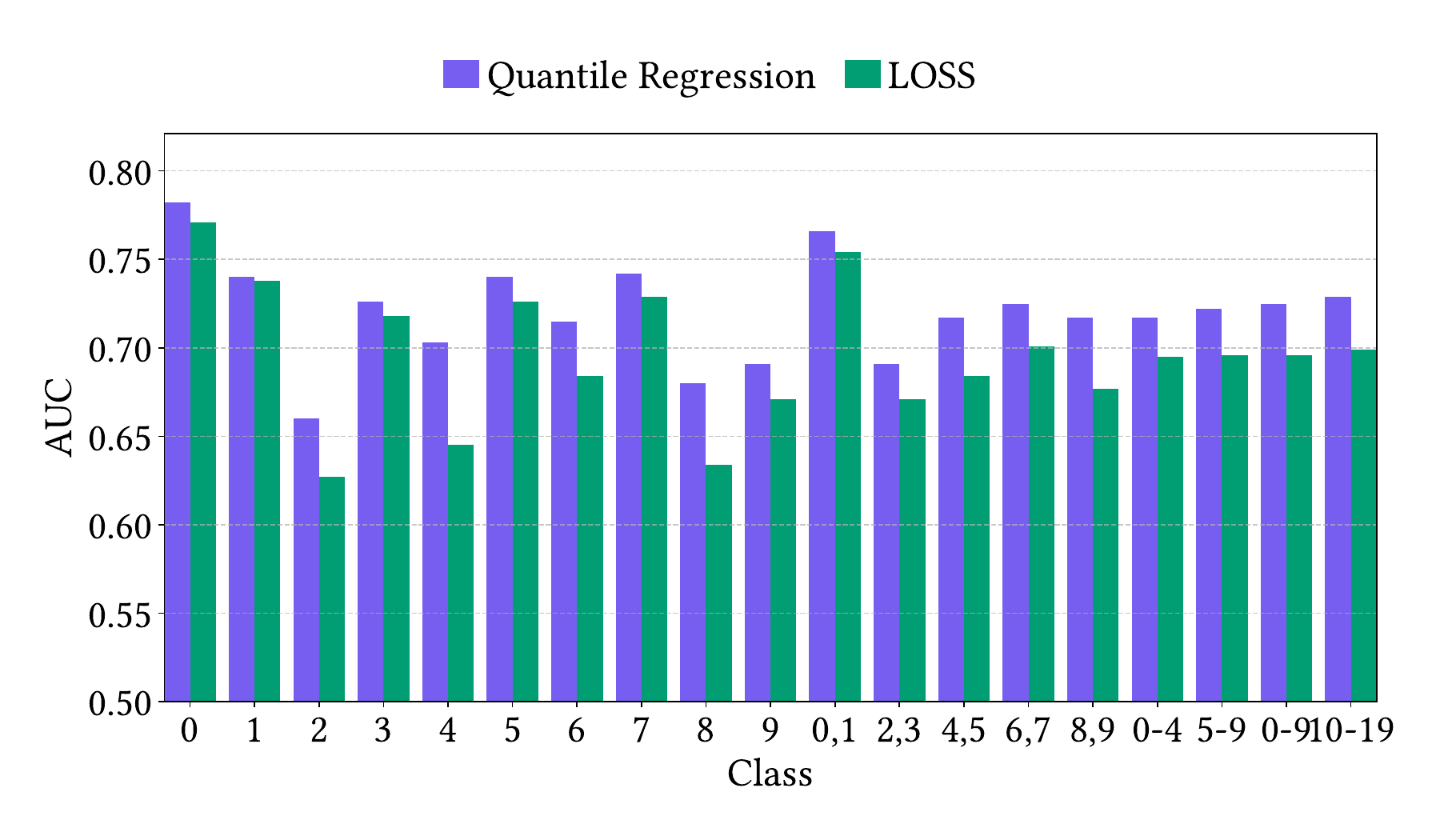} 
        \vspace{-.25in}
        \caption{AUC, CIFAR-100, ResNet-18.}
        \label{fig:qr_cifar100_auc_resnet18}
    \end{subfigure}
    \newline
    \begin{subfigure}[b]{0.5\textwidth}
        \centering
        \includegraphics[width=\textwidth]{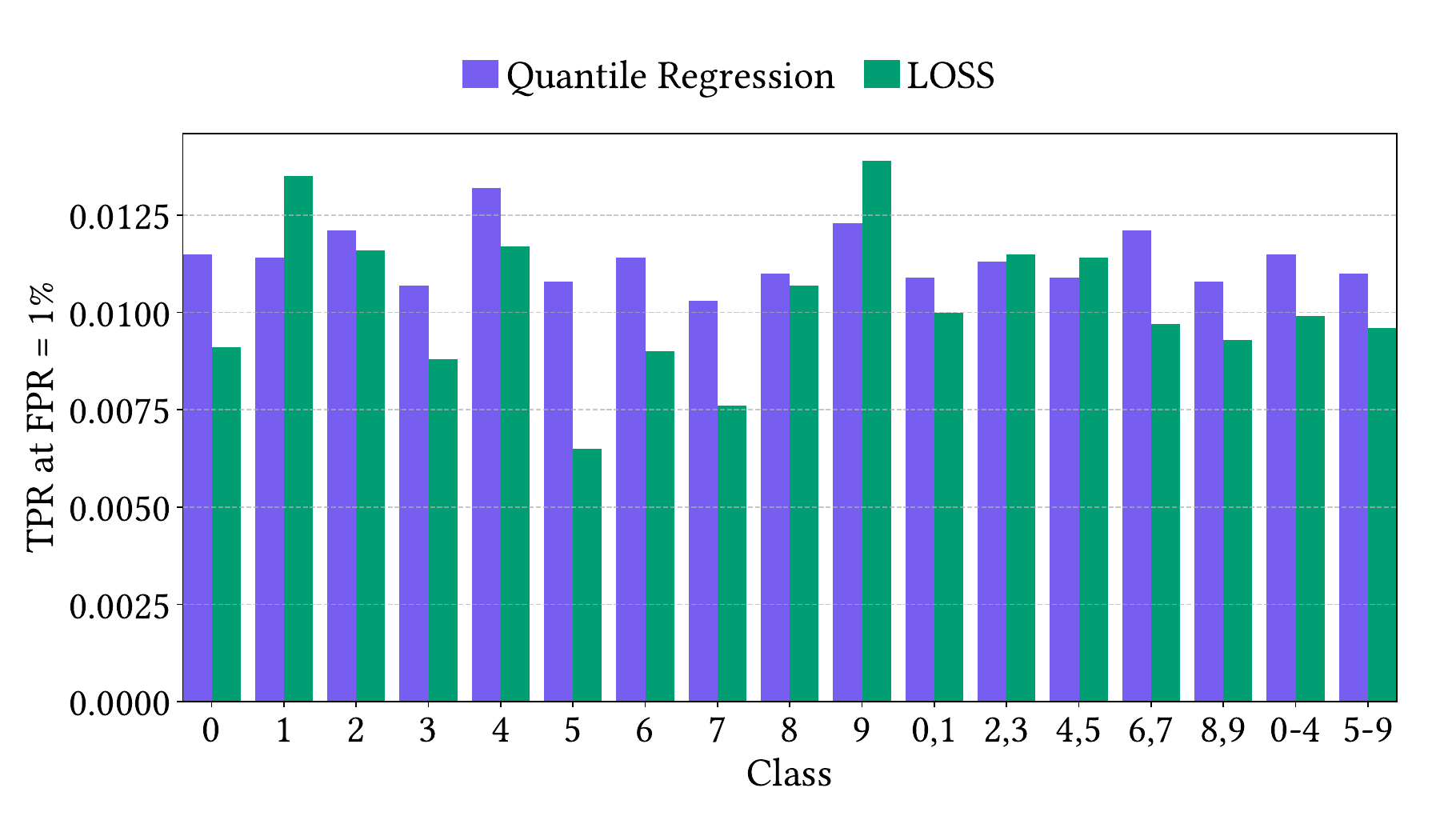} 
        \vspace{-.25in}
        \caption{TPR, CINIC-10, ViT.}
        \label{fig:appx:qr_cifar10_fpr1_vit}
    \end{subfigure}\hfill%
    \begin{subfigure}[b]{0.5\textwidth}
        \centering
        \includegraphics[width=\textwidth]{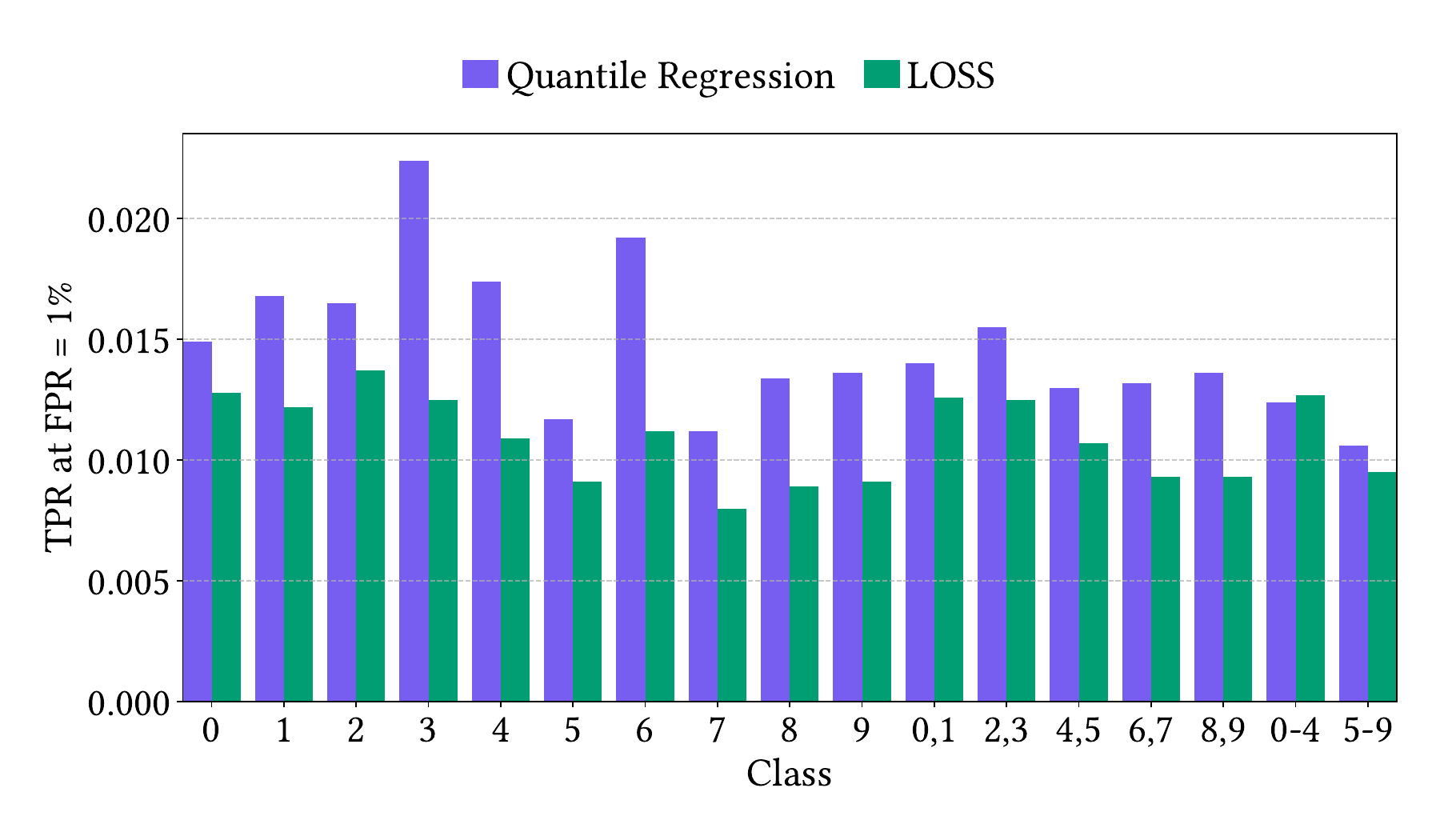} 
        \vspace{-.25in}
        \caption{TPR, CINIC-10, ResNet-18.}
        \label{fig:appx:qr_cifar10_fpr1_resnet18}
    \end{subfigure}
    \newline
    \begin{subfigure}[b]{0.5\textwidth}
        \centering
        \includegraphics[width=\textwidth]{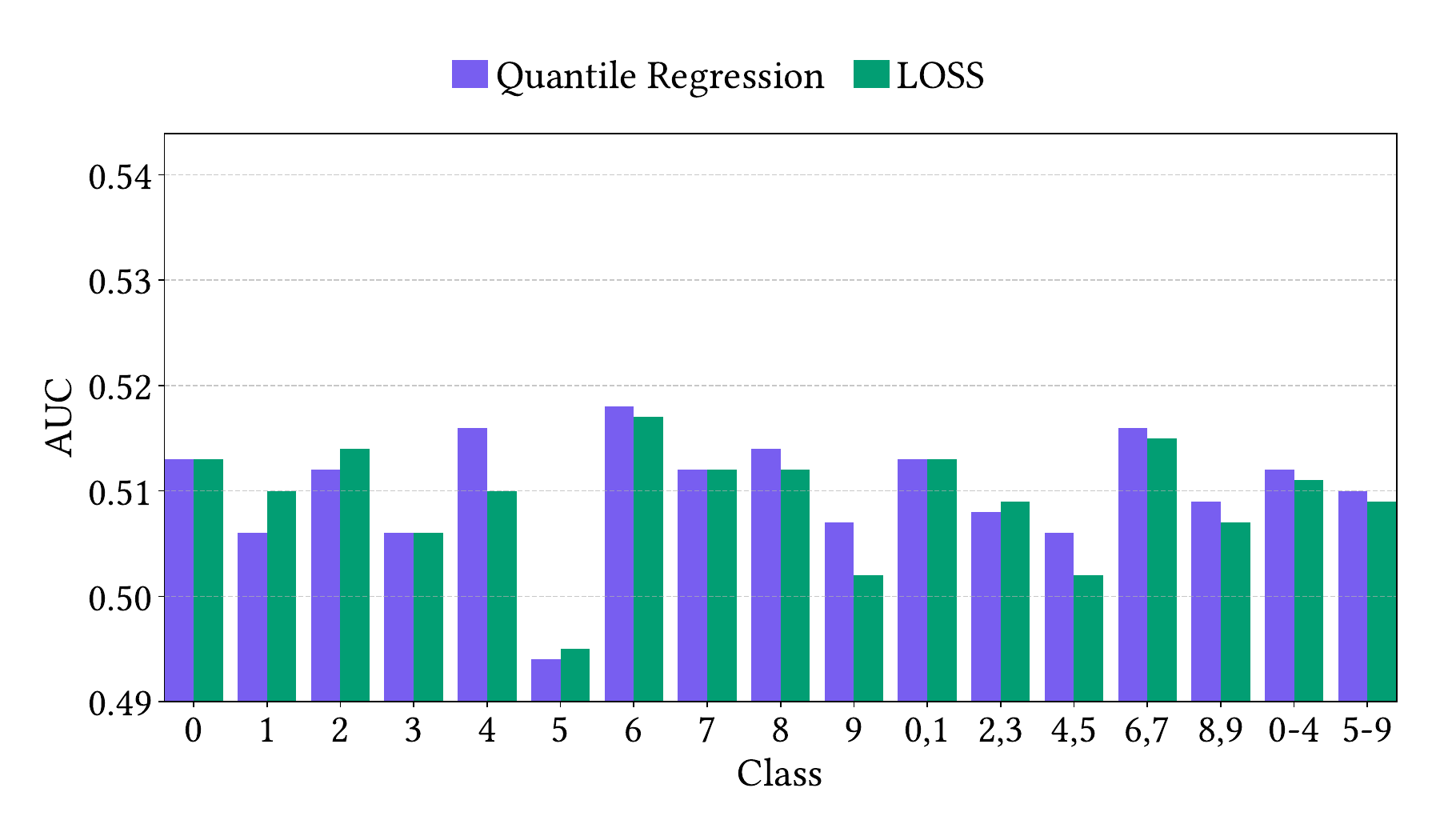} 
        \vspace{-.25in}
        \caption{AUC, CINIC-10, ViT.}
        \label{fig:qr_cifar10_auc_vit}
    \end{subfigure}\hfill%
    \begin{subfigure}[b]{0.5\textwidth}
        \centering
        \includegraphics[width=\textwidth]{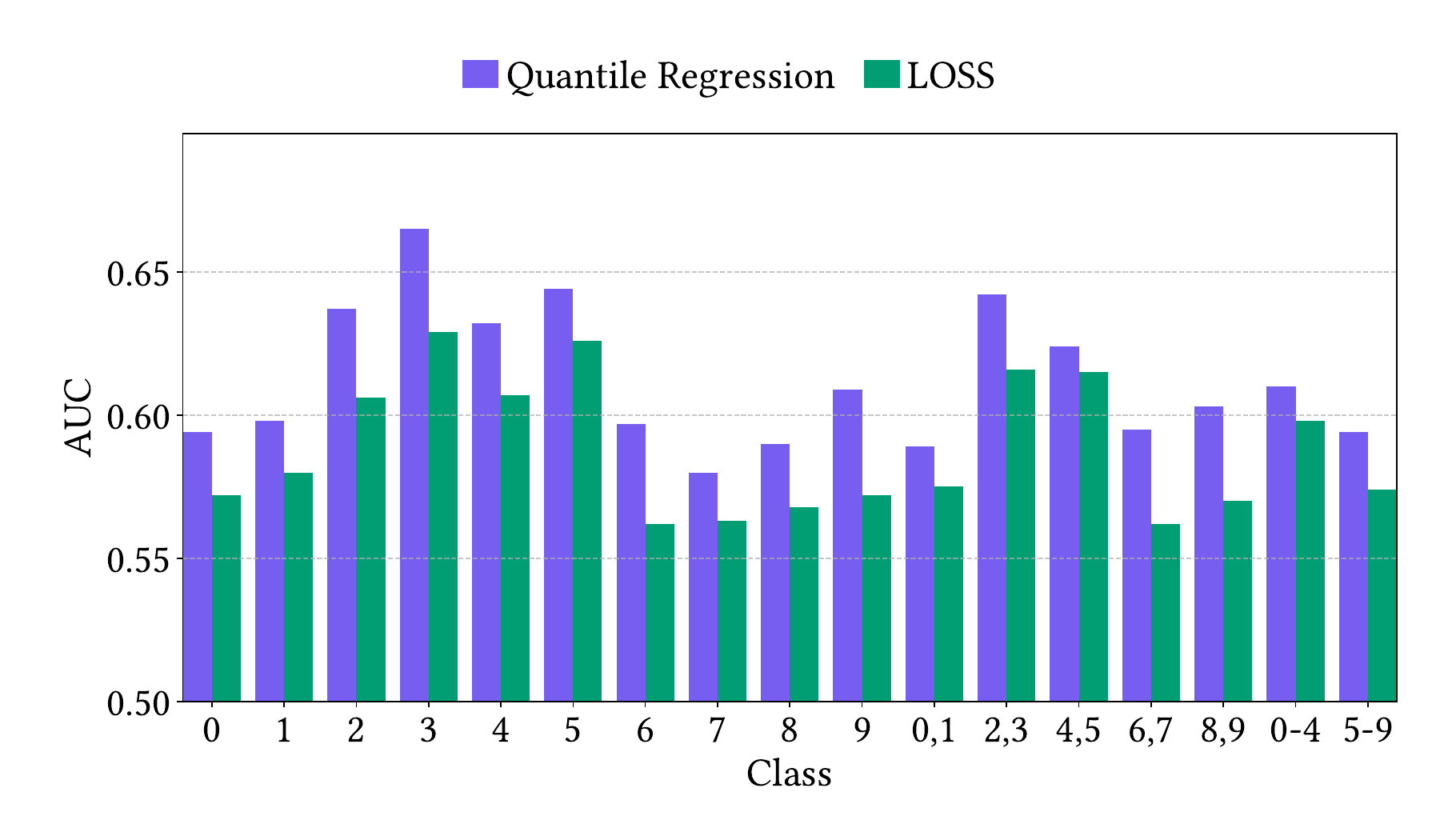} 
        \vspace{-.25in}
        \caption{AUC, CINIC-10, ResNet-18.}
        \label{fig:qr_cifar10_auc_resnet18}
    \end{subfigure}
    \caption{AUC and TPR for CINIC-10 and CIFAR-100 on sets of unseen classes. Each bar represents the metric on classes $C$ when $C$ are dropped from the attack training set. We only report results at 1\% FPR; the results at 0.1\% FPR are not meaningful due to the small sample size of the validation set on a single class.}
    \label{fig:appx:qr_alt_backbones}
\end{figure}

\clearpage
\section{Results on models trained with defenses}
\label{appx:defenses}

\paragraph{L2 Regularization.} For completeness, we study the unseen-class membership inference setting under common defenses. First, we measure the effect of L2 regularization, a known MIA defense, e.g., \citep{li2021membership}, \citep{choquette2021label}, \citep{leino2020stolen}, on our method. Since our models are trained with vanilla SGD, L2 regularization is equivalent to weight decay \cite{loshchilov2017decoupled}, so we proceed by modifying weight decay.

\begin{table}[h!]
\centering
\caption{Comparison of MIA on CIFAR-100 under Weight Decay}
\label{tab:combined_results}
\begin{tabular}{l ccc ccc}
\toprule
& \multicolumn{3}{c}{\textbf{TPR @ FPR = 1\%}} & \multicolumn{3}{c}{\textbf{AUC}} \\
\cmidrule(lr){2-4} \cmidrule(lr){5-7}
\textbf{Model} & \textbf{1cls} & \textbf{2cls} & \textbf{5cls} & \textbf{1cls} & \textbf{2cls} & \textbf{5cls} \\
\midrule
QMIA (wd = $5 \times 10^{-4}$)     & \textbf{4.59} & \textbf{3.38} & \textbf{3.04} & \textbf{0.709} & \textbf{0.726} & \textbf{0.686} \\
LOSS (wd = $5 \times 10^{-4}$) & 2.00 & 1.28 & 0.96 & 0.690 & 0.711 & 0.663 \\
\midrule
QMIA (wd = $5 \times 10^{-3}$)     & \textbf{2.50} & \textbf{1.73} & \textbf{1.95} & \textbf{0.576} & \textbf{0.591} & \textbf{0.612} \\
LOSS (wd = $5 \times 10^{-3}$) & 1.59 & 1.53 & 1.38 & 0.573 & 0.583 & 0.588 \\
\midrule
QMIA (wd = $5 \times 10^{-2}$)     & \textbf{1.25} & \textbf{1.16} & 0.81 & 0.520 & 0.505 & \textbf{0.512} \\
LOSS (wd = $5 \times 10^{-2}$) & 0.92 & 0.95 & \textbf{0.86} & \textbf{0.527} & \textbf{0.518} & \textbf{0.512} \\
\bottomrule
\end{tabular}
\end{table}

As L2 regularization (i.e. weight decay) increases, both QMIA and LOSS become less effective. However, the resulting model also has significantly lower test accuracy (76\% $\rightarrow$ 65\% $\rightarrow$ 55\%), so in general, sufficient L2 regularization is impractical.

\paragraph{Differential Privacy (DP).} We additionally analyze unseen-class membership inference under epsilon-delta DP guarantees; however, as shown in~\cite{leino2020stolen, lira, li2021membership, choquette2021label}, DP is an unrealistic defense when training models from scratch, as even modest guarantees require significantly degrading model performance.

In the below experiments, we train DP models from scratch on CIFAR-10 and CIFAR-100 with one class dropped out.

Following~\cite{leino2020stolen}, we evaluate DP at $\eps = 1.0, 4.0, 16.0$. Our training and test accuracies on CIFAR-10 and CIFAR-100 are comparable to the results in ~\cite{leino2020stolen}.
We find that DP in fact prevents quantile regression (as well as the LOSS baseline) from achieving meaningful TPR or AUCs in these settings.

\begin{table}[h!]
\centering
\caption{Performance Metrics Under Differential Privacy Settings for CIFAR10 and CIFAR100}
\label{tab:dp_performance}
\begin{tabular}{ll cc cc cc}
\toprule
& & & & \multicolumn{2}{c}{\textbf{TPR @ FPR = 1\%}} & \multicolumn{2}{c}{\textbf{AUC}} \\
\cmidrule(lr){5-6} \cmidrule(lr){7-8}
\textbf{Dataset} & $\boldsymbol{\epsilon}$ & \textbf{Train Acc.} & \textbf{Test Acc.} & \textbf{QMIA} & \textbf{LOSS} & \textbf{QMIA} & \textbf{LOSS} \\
\midrule
\multirow{3}{*}{CIFAR10} 
& 1  & \textbf{0.1005} & 0.0998 & 0.0056 & \textbf{0.0110} & 0.51 & 0.50 \\
& 4  & \textbf{0.1029} & 0.1007 & \textbf{0.0127} & 0.0123 & 0.52 & 0.52 \\
& 16 & 0.0975 & \textbf{0.1008} & 0.0076 & \textbf{0.0100} & 0.49 & 0.49 \\
\midrule
\multirow{3}{*}{CIFAR100} 
& 1  & 0.0522 & \textbf{0.0537} & 0.0025 & \textbf{0.0042} & 0.49 & 0.49 \\
& 4  & \textbf{0.0509} & 0.0498 & \textbf{0.0184} & 0.0067 & 0.51 & 0.51 \\
& 16 & 0.0494 & \textbf{0.0500} & 0.0025 & \textbf{0.0109} & 0.51 & 0.50 \\
\bottomrule
\end{tabular}
\end{table}

\clearpage
\section{Proof of Theorem~\ref{thm:transferability}}
\label{appx:proof}
\begin{proof}
As a first step, we prove by contradiction that the learned predictor $q^*_\alpha$ is $(\mathcal {W}, \phi, 0)$-multi-accurate under $P$. Suppose not, then, by the definition of multi-accuracy, there exists some $w' \in \mathcal{W}$ such that
\[
\mathbb{E}_{(x, s) \sim Q} \left[ \langle w', \phi(x) \rangle \cdot \left( \mathbf{1}\{s < q^*_\alpha(x)\} - \alpha \right) \right] \neq 0.
\]
Without loss of generality, suppose this expectation is strictly positive.

Since the pinball loss is convex and differentiable almost everywhere, its subgradient with respect to the weights at \( w^* \) is: 
\[
\nabla_w \mathbb{E}_{(x, s) \sim P}[\ell_\alpha(\langle w, \phi(x) \rangle, s)]\big|_{w = w^*} = - \mathbb{E}_{(x, s) \sim P} \left[ \left( \alpha - \mathbf{1}\{s < q^*_\alpha(x)\} \right) \phi(x) \right].
\]

Taking the inner product of this gradient with \( w' \), we obtain:
\[
\left\langle w', \nabla_w \mathbb{E}_{P}[\ell_\alpha(\langle w, \phi(x) \rangle, s)] \big|_{w = w^*} \right\rangle = - \mathbb{E}_{P} \left[ \langle w', \phi(x) \rangle \cdot \left( \alpha - \mathbf{1}\{s < q^*_\alpha(x)\} \right) \right] < 0,
\]
by assumption. Therefore, moving in the direction \( -w' \) decreases the expected pinball loss objective, contradicting the optimality of \( w^* \).

Thus, we must have:
\[
\left| \mathbb{E}_{(x, s) \sim P} \left[ \langle w, \phi(x)\rangle \cdot \left( \mathbf{1}\{s < q^*_\alpha(x)\} - \alpha \right) \right] \right| = 0 \quad \text{for all } w \in \mathcal{W},
\]
i.e., \( q^*_\alpha \) is $(\mathcal{W}, 0)$-multi-accurate under \( P \).

Finally, given that $\frac{dQ}{dP}(x)$ satisfies:
\[
\frac{dQ_\phi}{dP_\phi}(\phi(x),s) = \langle \phi(x), v \rangle \quad \text{for some } v \in \mathcal{W}, \text{ with } \langle \phi(x), v \rangle > 0 \text{ for all } x \in \text{supp}(Q),
\]
we can perform a change of measure from $P$ to $Q$:
\begin{align*}
 \mathbb{E}_{(x, s) \sim Q} \left[  \left( \mathbf{1}\{s < q^*_\alpha(x)\} - \alpha \right) \right] &=  \mathbb{E}_{(\phi(x), s) \sim Q_\phi} \left[  \left( \mathbf{1}\{s < \langle \phi(x), w^*\rangle\} - \alpha \right) \right] \\
 &= \mathbb{E}_{(\phi(x), s) \sim P_\phi} \left[ \frac{dQ_\phi(\phi(x), s)}{dP_\phi(\phi(x), s)} \left( \mathbf{1}\{s < \langle \phi(x), w^*\rangle\} - \alpha \right) \right]\\
&= \mathbb{E}_{(\phi(x), s) \sim P_\phi} \left[ \langle \phi(x), v\rangle \left( \mathbf{1}\{s < \langle \phi(x), w^*\rangle\} - \alpha \right) \right]\\
&= \mathbb{E}_{(x, s) \sim P} \left[ \langle \phi(x), v\rangle \left( \mathbf{1}\{s < q^*_\alpha(x)\} - \alpha \right) \right] = 0
\end{align*}

This completes the proof.
\end{proof}

\clearpage

\section{Enlarged Plots}
\label{appx:enlarged-plots}

\begin{figure}[h!]
    \centering
    \includegraphics[width=0.8\textwidth]{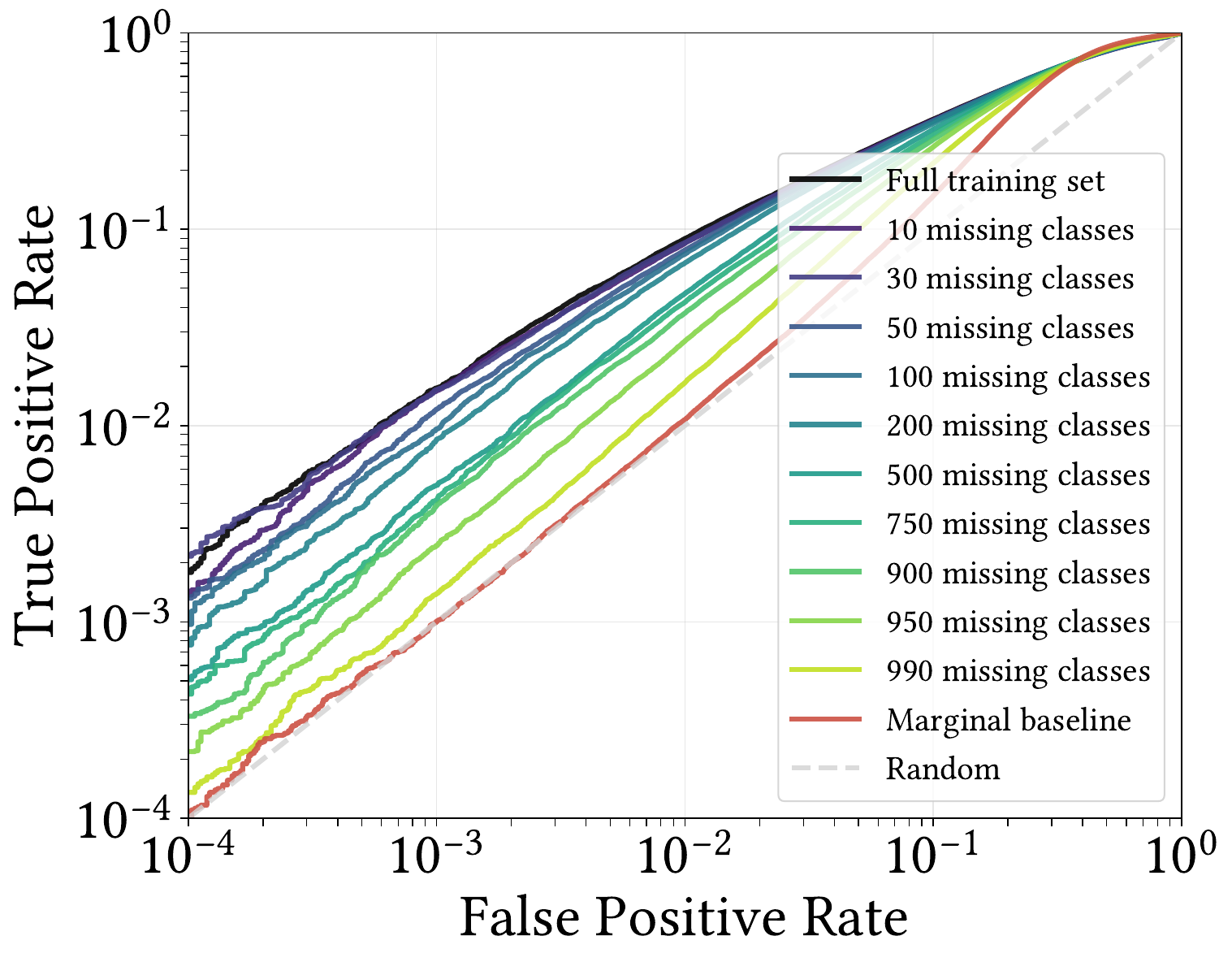}
    \caption{ROC curve for class drop experiment on ImageNet (Figure \ref{fig:imagenet_classdrop_roc} enlarged).}
    \label{fig:imagenet_classdrop_roc_large}
\end{figure}

\begin{figure}[h!]
    \centering
    \includegraphics[width=0.8\textwidth]{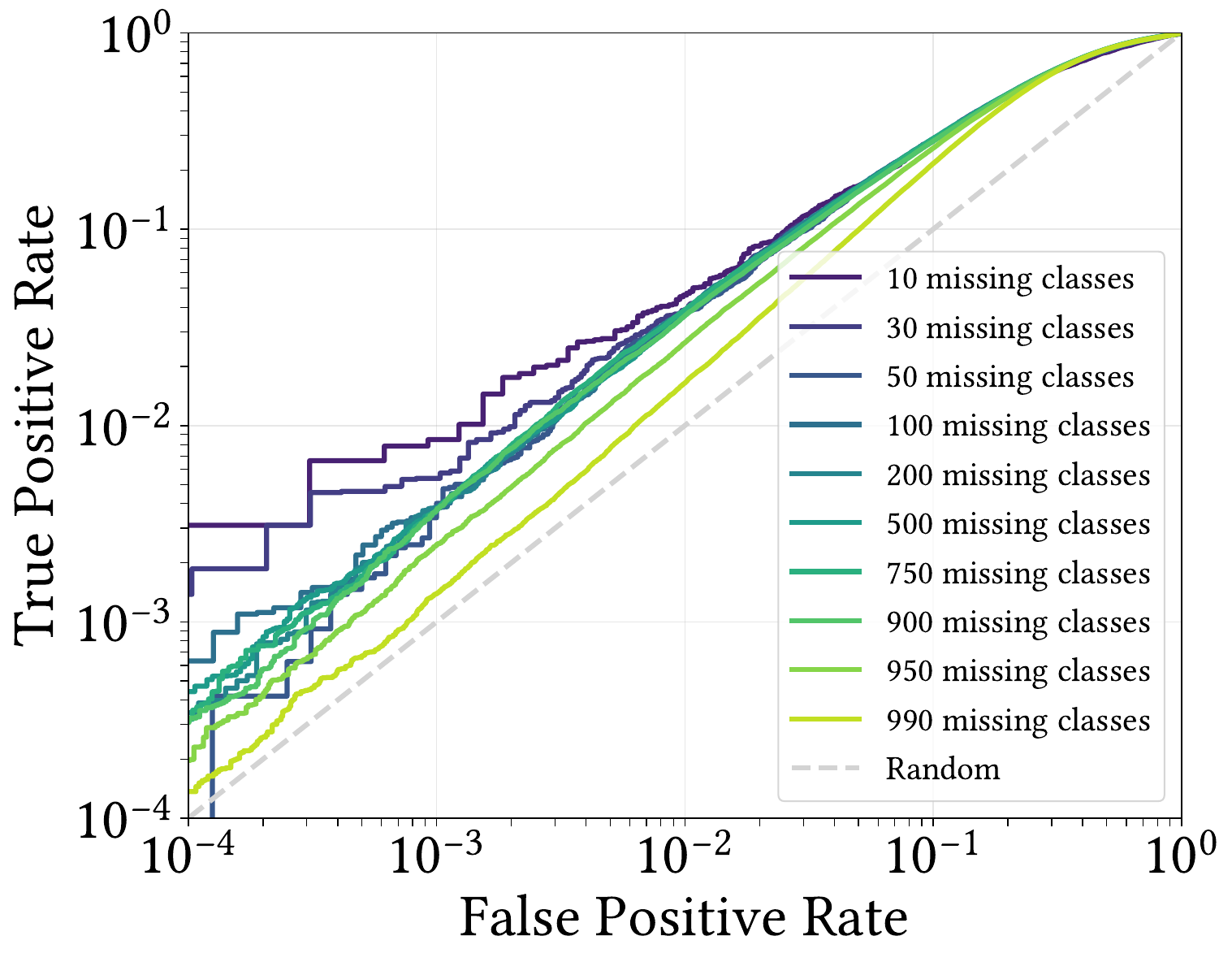}
    \caption{Unseen class ROC curve for class drop experiment on ImageNet (Figure \ref{fig:imagenet_classdrop_roc_ood} enlarged).}
    \label{fig:imagenet_classdrop_roc_ood_large}
\end{figure}

\begin{figure}[h!]
    \centering
    \includegraphics[width=0.8\textwidth]{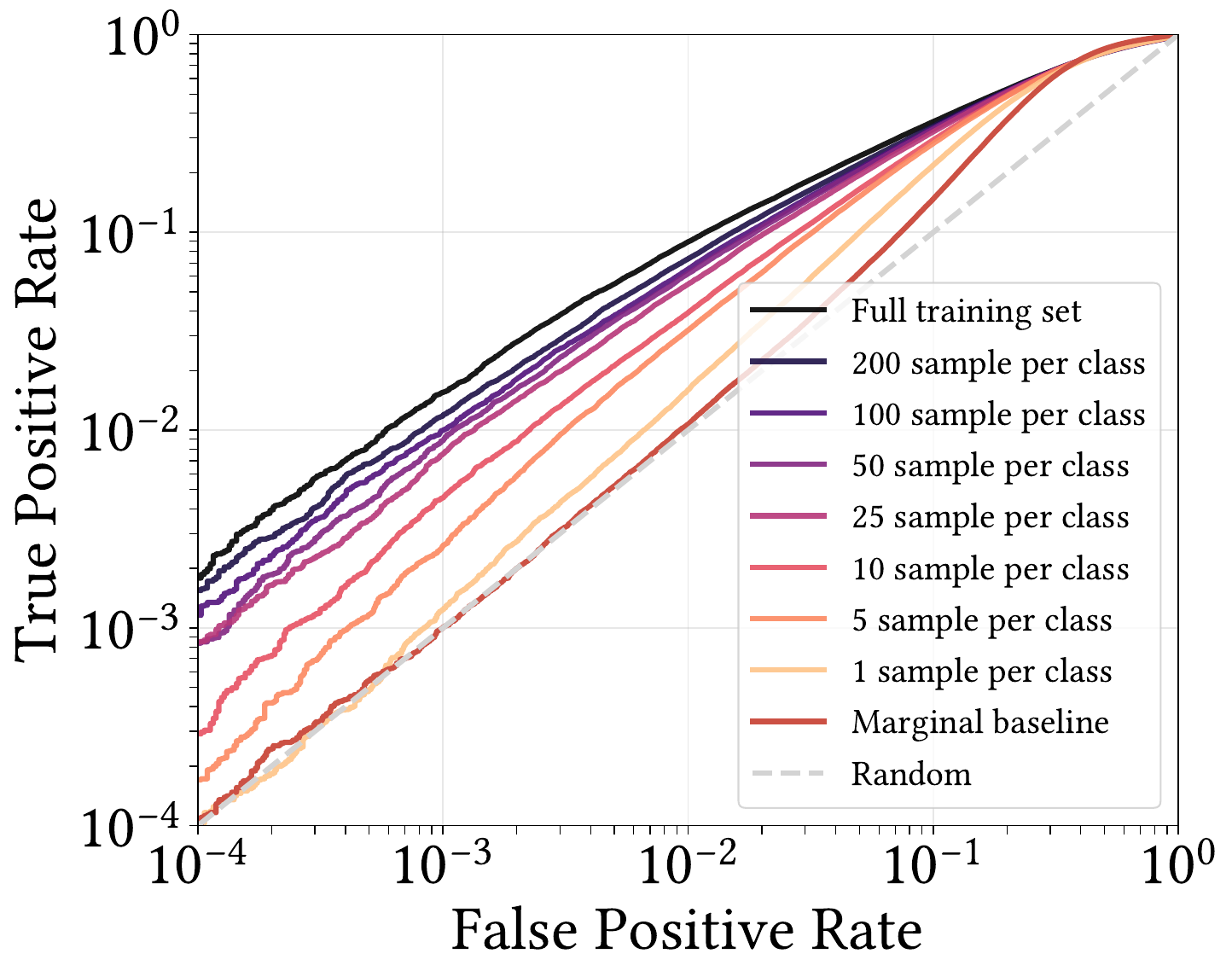}
    \caption{ROC curve for sample drop experiment on ImageNet. (Figure \ref{fig:imagenet_sampledrop_roc} enlarged).}
    \label{fig:imagenet_sampledrop_roc_large}
\end{figure}

\clearpage
\section{LLM Usage}

LLMs were used in writing boilerplate experiment code, debugging, and writing boilerplate plotting code. No LLMs were used for paper writing, research ideation, or related work discovery.

\end{document}